\documentclass[10pt,twocolumn,letterpaper]{article}

\ifdefined\cvprsubmit
  \usepackage[review]{cvpr}
\else
  \usepackage[pagenumbers]{cvpr}
\fi

\usepackage{multirow}
\usepackage{colortbl}
\usepackage{makecell}
\usepackage{xspace}
\usepackage{siunitx}
\usepackage{pifont}
\usepackage{tikz}
\usetikzlibrary{arrows.meta,positioning,fit,backgrounds,calc}
\usepackage{placeins}   %
\usepackage{algorithm}
\usepackage[noend]{algpseudocode}
\algrenewcommand\algorithmicrequire{\textbf{Input:}}
\algrenewcommand\algorithmiccomment[1]{\hfill\mbox{\hypersetup{allcolors=cInkTwo}\color{cInkTwo}\(\triangleright\)~#1\hspace{0.7em}}}

\usepackage[protrusion=true,expansion=true,final]{microtype}
\makeatletter
\renewcommand\paragraph{\@startsection{paragraph}{4}{\z@}%
  {6pt \@plus 2pt \@minus 1pt}{-0.8em}{\normalfont\normalsize\bfseries}}
\makeatother

\newcommand{\tablestyle}[1][5pt]{\small\setlength{\tabcolsep}{#1}\renewcommand{\arraystretch}{1.2}}
\newcommand{\tablenote}{\footnotesize\setlength{\baselineskip}{10.5pt}}  %
\newcommand{\blockgap}{\addlinespace[3pt]}                                 %
\newsavebox{\tabmainbox}                                                  %
\newcommand{\subrow}{\hspace*{1.1em}}                                       %
\newcommand{\oursrow}{\rowcolor{cOurs!12}}
\AtEndPreamble{%
  \crefname{algorithm}{Algorithm}{Algorithms}\Crefname{algorithm}{Algorithm}{Algorithms}%
  \crefname{figure}{Fig.}{Figs.}\Crefname{figure}{Figure}{Figures}%
  \crefname{equation}{Eq.}{Eqs.}\Crefname{equation}{Equation}{Equations}%
  \creflabelformat{equation}{#2#1#3}}   %

\ifdefined\cvprsubmit
  \newcommand{\Suppname}{The supplementary material\xspace}
\else
  \newcommand{\Suppname}{The appendix\xspace}
\fi
\ifdefined\cvprsuppl
  \newcommand{\mainref}[1]{\cref*{#1}}
  \newcommand{\Mainref}[1]{\Cref*{#1}}
\else
  \newcommand{\mainref}[1]{\cref{#1}}
  \newcommand{\Mainref}[1]{\Cref{#1}}
\fi

\ifdefined\cvprsubmit
  
  \newcommand{\reviewlastpage}{}   %
\else
  
  \newcommand{\reviewlastpage}{}
\fi

\newcommand{\cexec}{\mathrm{exec}}
\newcommand{\ctool}{\mathrm{tool}}
\newcommand{\rarr}{r_{\mathrm{arr}}}
\usepackage{dsfont}
\newcommand{\ind}{\mathds{1}}

\newcommand{\method}{\textsc{H2E}\xspace}          %

\newcommand{\reverie}{REVERIE-CE\xspace}

\newcommand{\cmark}{\ding{51}}
\newcommand{\xmark}{\ding{55}}

\newcommand{\best}[1]{\textbf{#1}}

\newcommand{\gray}[1]{{\color{black!62}#1}}   %

\definecolor{cBase}{HTML}{2A78D6}     %
\definecolor{cAdapt}{HTML}{1BAF7A}    %
\definecolor{cOurs}{HTML}{EB6834}     %
\definecolor{cInk}{HTML}{0B0B0B}
\definecolor{cInkTwo}{HTML}{52514E}
\definecolor{cGrid}{HTML}{E4E3DF}
\definecolor{cPanel}{HTML}{F4F3EF}

\ifdefined\cvprsubmit
  \allowdisplaybreaks[1]   %
\fi
\ifdefined\cvprsuppl
  \usepackage{xr-hyper}
\fi

\definecolor{cvprblue}{rgb}{0.21,0.49,0.74}
\usepackage[pagebackref,breaklinks,colorlinks,allcolors=cvprblue]{hyperref}
\hypersetup{pdftitle={What Can a VLA Learn from the Agent That Calls It? Task-Level Experience Distillation for Embodied Navigation}}
\ifdefined\cvprsubmit
  \hypersetup{pdfauthor={}, pdfsubject={}, pdfkeywords={}}
\else
  \hypersetup{pdfauthor={Jinze Tao, Yuxiao Han, Kangkang Dong, Yancong Wei, Xiaojun Zhu, Bokui Chen, Houde Liu}}
\fi

\title{What Can a VLA Learn from the Agent That Calls It?\\
Task-Level Experience Distillation for Embodied Navigation}

\newcommand{\arxivauthors}{%
Jinze Tao\quad\ \ Yuxiao Han\quad\ \ Kangkang Dong\quad\ \ Yancong Wei\\[0.15em]
Xiaojun Zhu$^{*}$\quad\ \ Bokui Chen\quad\ \ Houde Liu\\[0.6em]
{\normalsize Tsinghua Shenzhen International Graduate School, Tsinghua University}\\[0.15em]
{\small\tt tao.jinze@163.com\qquad zhu.xj@foxmail.com}}
\ifdefined\cvprsubmit
\author{Anonymous}   %
\else
\author{\arxivauthors}
\fi

\begin{document}
\ifdefined\cvprsuppl
\clearpage
\ifdefined\cvprsuppl
  \renewcommand{\thetable}{S\arabic{table}}
  \renewcommand{\thefigure}{S\arabic{figure}}
  \renewcommand{\theequation}{S\arabic{equation}}
\fi
\makeatletter
\providecommand{\citenoback}[1]{{\let\Hy@backout\@gobble\cite{#1}}}
\makeatother
\twocolumn[{%
  \centering
  \ifdefined\cvprsuppl
    {\Large\textbf{\thetitle}\\[0.5em]\Large Supplementary Material\\[1.2em]}
  \else
    {\Large\textbf{Appendix}\\[0.4em]\large\thetitle\\[1.2em]}
  \fi
  \tablestyle[26pt]
  \sbox{\tabmainbox}{%
  \begin{tabular}{@{\hspace{\tabcolsep}}lcccc@{\hspace{\tabcolsep}}}
    \toprule
    Model & SR$\uparrow$ & OSR$\uparrow$ & SPL$\uparrow$ & NE$\downarrow$ \\
    \midrule
    Uni-NaVid~\citenoback{uninavid} & 21.6 & 48.7 & 9.3 & 10.18 \\
    \oursrow
    \;\;+ \method             & \best{30.9} & \best{51.5} & \best{18.9} & \best{9.02} \\
    \bottomrule
  \end{tabular}}%
  \captionsetup{type=table,width=\wd\tabmainbox}
  \caption{\textbf{Uni-NaVid with and without \method on \reverie val-unseen.}}
  \label{tab:supp-unv}
  \usebox{\tabmainbox}
  \vspace{0.8em}
}]
\appendix

\section{\method on other VLA}
\label{sec:supp-secondvla}
The three rules of \mainref{sec:labels} turn successful agent sessions into supervision for a VLA.
They decide which steps to learn from, under which instruction, and where the task ends. These
choices concern the task and not the architecture of the VLA, so we keep the three rules and adapt
only the observations and the action targets. Uni-NaVid~\citenoback{uninavid} is a 7B video VLA that
reads a monocular forward video and emits discrete actions.

The motion of a session is its recorded pose sequence $p_t$, which does not depend on how a VLA
represents its actions. A VLA with a discrete action head cannot learn from these poses directly. A
converter therefore turns each pose sequence into the actions of that VLA and renders its
observations at the poses that the actions reach.

\begin{figure*}[t]
  \centering
  \includegraphics[width=\textwidth]{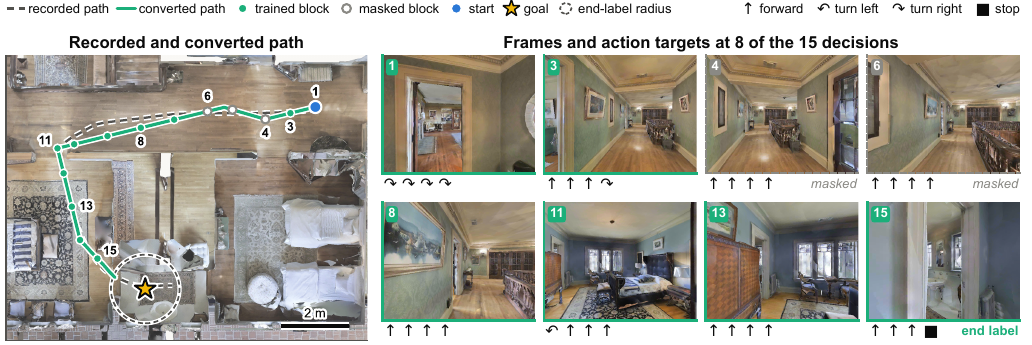}
  \caption{\textbf{Converting one recorded session for Uni-NaVid.} The converted path follows the recorded one with
  discrete actions. Each dot is a decision state with one frame and a block of four action targets, and hollow dots
  mark blocks that progress masking excludes from the loss.}
  \label{fig:supp-unv-conv}
\end{figure*}

\begin{figure*}[t]
  \centering
  \includegraphics[width=\textwidth]{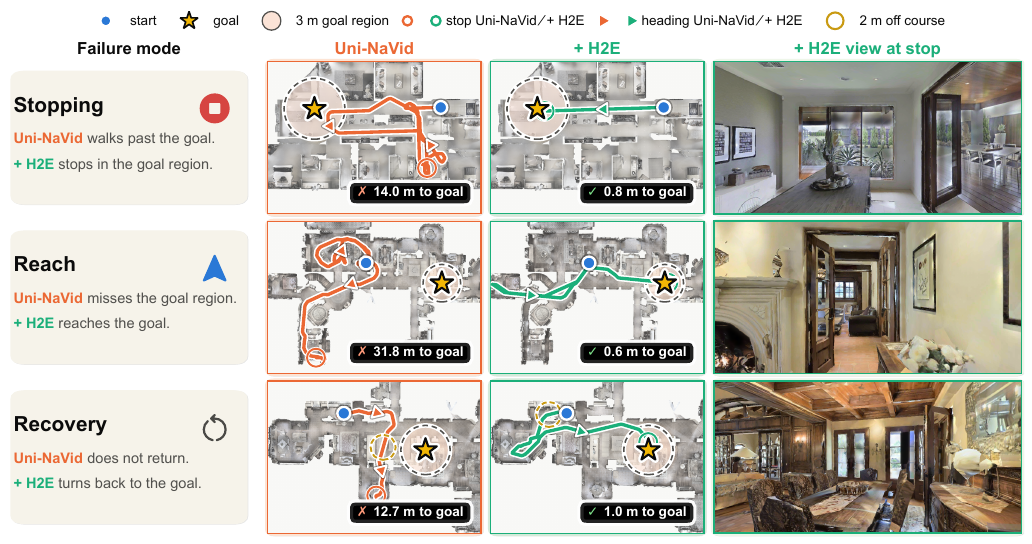}
  \caption{\textbf{Stopping, reach and recovery of Uni-NaVid with and without \method.} Each row pairs the two paths
  on one val-unseen episode of \reverie, with the view at the stop of \method and the final distance
  to the goal.}
  \label{fig:supp-unv-qual}
\end{figure*}

\paragraph{Discrete actions.}
Uni-NaVid chooses among forward, turn left, turn right and stop. Its executor moves 0.25\,m or turns
$30^\circ$, and the converter drives this executor in the simulator. Let $\gamma(s)$ be the recorded
path at travelled length $s$, $S$ its length, and $s_k$ the progress of the converted pose
$\tilde{p}_k$ along it. We take $s_k$ as the length within 1.5\,m beyond $s_{k-1}$ whose path point
lies closest to $\tilde{p}_k$, so the progress never moves back. With a look-ahead of $L=1$\,m,
\begin{equation}
  \begin{aligned}
    u_k &= \gamma\big(\min\{s_k+L,\,S\}\big),\\
    a_k &= \kappa(\tilde{p}_k,\,u_k),\qquad \tilde{p}_{k+1} = F(\tilde{p}_k,\,a_k).
  \end{aligned}
  \label{eq:disc}
\end{equation}
Here $k$ counts controller updates, and $a_k$ is one forward step or a run of turns. $\kappa$ aims at
the first corner of the navmesh shortest path to $u_k$. When the heading error reaches $22.5^\circ$ it
outputs the number of $30^\circ$ turns closest to the error, and otherwise one forward step. The
threshold drops to $15^\circ$ after four consecutive updates without 5\,cm of new progress. The
converter discards a conversion after more than 40 updates without such progress. $F$
executes $a_k$ with the executor of the evaluation environment, so every label is an action that the
VLA can carry out at test time. Aiming ahead keeps the converted path close to the recorded one. For
each training session we take the largest distance from a converted pose to the nearest recorded
pose, and the median over sessions is 0.32\,m.

\paragraph{Observations.}
Discrete execution changes the poses, so the converter renders the forward camera at the poses it
reaches. It uses the camera settings of the evaluation environment and resizes the frames to the input
size of the VLA. Uncompressed renders match the frames of the evaluation environment pixel for pixel
at equal poses. A Uni-NaVid call adds one frame and returns four actions. The converter therefore
lists every turn of a run as its own action, cuts the action stream into blocks of four, and renders
one frame before each block. A sample sees the frames up to its own.

\paragraph{Labels.}
Task-level labels of Eq.~\ref{eq:rc} give every sample the original instruction, whichever actor moved
the robot. The target of a sample is its block of four actions, padded with stop at the end. Progress
masking of Eq.~\ref{eq:mask} works on these blocks. A block enters the loss only if it passes one of two
tests. In the first, one of its states meets Eq.~\ref{eq:term}, counting the state after its last action.
In the second, its last state is at most $\delta$ farther from the goal than its first. A masked block keeps its frame in the video. The end
label of Eq.~\ref{eq:term} becomes the stop action, and the converted stream ends there. If the converted
pose reaches the end of the recorded path without meeting Eq.~\ref{eq:term}, the controller follows the
geodesic path to the goal instead of $u_k$. Only conversions that reach the end label give training samples. Figure~\ref{fig:supp-unv-conv} shows a
converted session. Three of its blocks walk away from the goal and are masked, and its last block ends
with the end label.

\paragraph{Training.}
Uni-NaVid trains on the converted agent sessions, DART and R2R-CE. It keeps the mixture weights, the
LoRA rank, the scale, the learning rate, the schedule and the number of updates of
\mainref{tab:setup}. Only the LoRA weights train. We replace the $L_1$ waypoint loss of Eq.~\ref{eq:loss} with the
negative log-likelihood of the action tokens. Each block gives a sample $x_j=(o_{\le j},\ell,y_j,m_j)$,
where $y_j$ is the target token sequence and $m_j$ the mask. The loss is
\begin{equation}
  \mathcal{L}(\theta) = \sum_{s} w_s\,\mathbb{E}_{x_j\sim\mathcal{D}_s}
  \big[-m_j \log \pi_\theta\big(y_j \mid o_{\le j},\ell\big)\big].
  \label{eq:loss-ce}
\end{equation}
We evaluate the final checkpoint. Each call adds the newest frame to the video of the episode and
returns four actions. The robot executes them in order, and a stop ends the episode. Uni-NaVid runs under the step and time limits of Sec.~\ref{sec:supp-impl},
and an episode succeeds when the robot stops within 3\,m of the goal.

\begin{figure}[t]
  \centering
  \includegraphics[width=\linewidth]{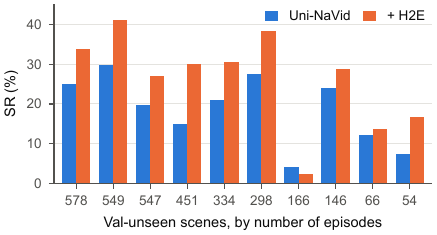}
  \caption{\textbf{SR of Uni-NaVid on each val-unseen scene} with and without \method. Scenes are
  ordered by their number of episodes, shown on the axis.}
  \label{fig:supp-unv-scenes}
\end{figure}

\begin{table}[t]
  \centering
  \tablestyle[5pt]
  \caption{\textbf{Stopping, recovery and path efficiency of Uni-NaVid} on \reverie val-unseen, in
  percent.}
  \label{tab:supp-unv-skill}
  \begin{tabular}{@{\hspace{\tabcolsep}}lcccc@{\hspace{\tabcolsep}}}
    \toprule
    Model & $C\uparrow$ & Stop prec.$\uparrow$ & Recovery$\uparrow$ & Path eff.$\uparrow$ \\
    \midrule
    Uni-NaVid & 44.4 & 29.0 & 27.1 & 43.0 \\
    \oursrow
    \;\;+ \method & \best{60.0} & \best{45.3} & \best{30.4} & \best{61.3} \\
    \bottomrule
  \end{tabular}
\end{table}

\paragraph{Results.}
Table~\ref{tab:supp-unv} compares the released Uni-NaVid weights with the same model trained with
\method. Both run on all \num{3189} val-unseen episodes of \reverie with the same evaluation code and
seed. \method raises the SR of Uni-NaVid by 9.3\%, to 30.9\%, and more than doubles its SPL, from
9.3\% to 18.9\%. It succeeds on 663 episodes on which the released weights fail. Figure~\ref{fig:supp-unv-qual} shows three of them, one per skill, drawn as
in \mainref{fig:qual}. Figure~\ref{fig:supp-unv-scenes} splits SR by scene. The gain holds in nine
of the ten val-unseen scenes, including the six largest.

\paragraph{Where the gain comes from.}
We measure stopping, reaching and recovery as in \mainref{sec:exp-mech}. Table~\ref{tab:supp-unv-skill}
reports stopping, recovery and path efficiency, and Tab.~\ref{tab:supp-unv} reports OSR for reaching. All
of them improve. For stopping, $C$ is the share of goal-region visits that end in success, and it
rises from 44.4\% to 60.0\%. The stop precision is the share of
executed stops that land inside the goal region, and it rises from 29.0\% to 45.3\%. For reaching, OSR
rises from 48.7\% to 51.5\%. For recovery we use the off-track rule of
\mainref{sec:exp-mech}, and recovery is the share of these episodes that later reach the goal region. It
rises from 27.1\% to 30.4\%. Path efficiency is the mean SPL of successful episodes. It grows from
43.0\% to 61.3\%, so successful episodes follow paths much closer to the shortest path. Uni-NaVid
thus gains the same three skills as the waypoint VLA. Together with \mainref{tab:main}, these results show that task-level experience of an
LLM agent improves both a waypoint VLA and a discrete-action VLA.

\section{More results}
\label{sec:supp-results}
\begin{table}[t]
  \centering
  \tablestyle[3.2pt]
  \caption{\textbf{Three benchmarks on val-unseen.}}
  \label{tab:gen}
  \begin{tabular}{@{}llcccc@{}}
    \toprule
    Benchmark & Method & SR$\uparrow$ & OSR$\uparrow$ & SPL$\uparrow$ & NE$\downarrow$ \\
    \midrule
    \reverie  & Base VLA              & 35.0 & 52.0 & 26.5 & 8.53 \\
              & Geodesic adaptation   & 34.9 & 38.5 & 33.2 & 8.93 \\
              & \method, one round    & 44.9 & 50.1 & 38.4 & 6.90 \\
    \oursrow
              & \method, two rounds   & \best{50.6} & \best{56.3} & \best{40.8} & \best{6.26} \\
    \midrule
    NavRAG-CE & Base VLA              & 21.3 & 60.8 & 14.1 & 9.41 \\
              & \method, one round    & 31.8 & 59.4 & \best{23.8} & 7.94 \\
    \oursrow
              & \method, two rounds   & \best{32.4} & \best{62.6} & 20.6 & \best{7.80} \\
    \midrule
    R2R-CE    & Base VLA              & \best{76.9} & \best{82.3} & 70.5 & \best{2.94} \\
              & \method, one round    & 74.1 & 76.2 & \best{70.7} & 3.42 \\
    \bottomrule
  \end{tabular}
\end{table}

\paragraph{Three benchmarks.}
Table~\ref{tab:gen} reports \method and the base VLA on all three val-unseen splits. \method keeps a high
success rate of 74.1\% on R2R-CE and gains on the two goal-directed splits. On NavRAG-CE, the second round
adds 0.6\% SR, 3.2\% OSR and lowers NE by 0.14\,m over the first round.

\begin{table*}[t]
  \centering
  \tablestyle[6pt]
  \caption{\textbf{Outcomes of the models of \mainref{tab:recipe} on the 100 held-out episodes.} Late means the robot
  reached the 3\,m region and ended outside it. Early means it stopped without reaching it. Cut means the run
  ended without a stop, and one run in the first and one in the third row count as both Late and Cut. Rec.\ counts
  recoveries after going off track.}
  \label{tab:supp-dev}
  \begin{tabular}{lcccccccc}
    \toprule
    Model & SR & OSR & $C$ & Late & Early & Cut & Only-H & Rec. \\
    \midrule
    Base VLA            & 38 & 59 & 0.64 & 21 & 38 & 4 & 0 & 5/39 \\
    Geodesic adaptation & 36 & 43 & 0.84 & 7  & 55 & 2 & 1 & 0/45 \\
    \oursrow
    \method      & 53 & 57 & 0.93 & 4  & 41 & 3 & 8 & 6/40 \\
    \midrule
    Agent with base VLA & 53 & 70 & --   & -- & -- & -- & 12 & -- \\
    \bottomrule
  \end{tabular}
\end{table*}
\paragraph{Held-out episodes.}
The 100 episodes of \mainref{tab:harness} come from 26 training-split scenes that the models of these tables never train on. Each
episode has its own target object. Table~\ref{tab:supp-dev} splits the outcomes of the models of \mainref{tab:recipe}
on these episodes. Only-H counts the 12 episodes that only the agent reached.

\method reaches 8 of the 12 goals that only the agent found, and geodesic adaptation reaches 1. \method
therefore transfers part of the goal finding of the agent to the VLA.

\begin{figure}[t]
  \centering
  \includegraphics[width=\linewidth]{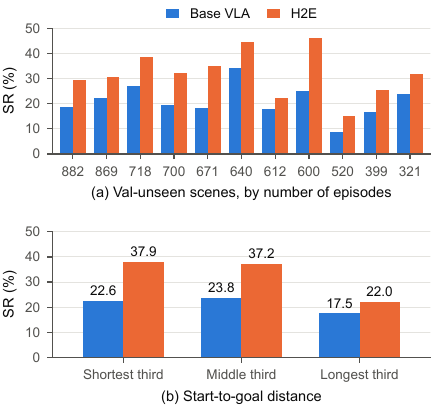}
  \caption{\textbf{SR on NavRAG-CE val-unseen} by scene and by thirds of the start-to-goal distance. Scenes are
  ordered by their number of episodes, shown on the axis.}
  \label{fig:supp-nrag}
\end{figure}
\begin{table}[t]
  \centering
  \tablestyle[5pt]
  \caption{\textbf{SR by instruction length} on val-unseen, in \%. Each benchmark is split into three groups by
  the number of words.}
  \label{tab:supp-len}
  \begin{tabular}{@{\hspace{\tabcolsep}}llccc@{\hspace{\tabcolsep}}}
    \toprule
    Benchmark & Words & Episodes & Base VLA & \method \\
    \midrule
    \reverie  & 14 or fewer & \num{1035} & 33.1 & \best{48.5} \\
              & 15 to 19    & \num{1086} & 38.0 & \best{52.7} \\
              & 20 or more  & \num{1068} & 33.6 & \best{50.5} \\
    \midrule
    NavRAG-CE & 20 or fewer & \num{2104} & 18.8 & \best{28.9} \\
              & 21 to 24    & \num{2089} & 20.9 & \best{30.8} \\
              & 25 or more  & \num{2739} & 23.6 & \best{36.2} \\
    \bottomrule
  \end{tabular}
\end{table}

\paragraph{Where the gain holds.}
Figure~\ref{fig:supp-nrag} splits SR on NavRAG-CE, the benchmark that \method never trains on, as
\mainref{fig:analysis} does for \reverie. Two rounds of \method raise SR in all 11 val-unseen scenes and
in every third of the episodes by start-to-goal distance.

Table~\ref{tab:supp-len} splits both benchmarks by the length of the instruction. \method raises SR in all
three groups on both benchmarks. The smallest gains are 14.6\% on \reverie and 9.9\% on NavRAG-CE.

On the episodes that both models solve, \method takes fewer steps. The median falls from 58 to 46 steps
on the 785 such episodes of \reverie and from 44 to 37 steps on the 944 such episodes of NavRAG-CE.

\begin{table}[t]
  \centering
  \tablestyle[4.2pt]
  \caption{\textbf{Three skills on both val-unseen benchmarks.} Each cell is a gain over the base VLA in \%.}
  \label{tab:supp-cap}
  \begin{tabular}{@{\hspace{\tabcolsep}}lcccc@{\hspace{\tabcolsep}}}
    \toprule
    Gain over the base VLA & $\Delta$SR & Reach & Stop & Recovery \\
    \midrule
    \multicolumn{5}{@{\hspace{\tabcolsep}}l}{\textit{\reverie, \num{3189} episodes}} \\
    \subrow \method, one round  & $+9.9$         & $-1.7$        & $+11.6$ & $+2.2$ \\
    \oursrow
    \subrow \method, two rounds & \best{$+15.6$} & \best{$+3.8$} & \best{$+11.8$} & \best{$+7.4$} \\
    \midrule
    \multicolumn{5}{@{\hspace{\tabcolsep}}l}{\textit{NavRAG-CE, \num{6932} episodes}} \\
    \subrow \method, one round  & $+10.5$        & $-0.7$        & \best{$+11.2$} & $+2.8$ \\
    \oursrow
    \subrow \method, two rounds & \best{$+11.1$} & \best{$+0.9$} & $+10.1$ & \best{$+6.9$} \\
    \bottomrule
  \end{tabular}
\end{table}

\paragraph{Three skills on both benchmarks.}
We apply the decomposition of \mainref{eq:decomp} to all \num{3189} val-unseen episodes of
\reverie and to all \num{6932} val-unseen episodes of NavRAG-CE. For a candidate and the base VLA
we read SR and OSR from the same episode runs, which gives the reach term and the stop term as a
share of all episodes. For recovery we use the off-track rule of \mainref{sec:exp-mech}. An
episode goes off track when its per-step geodesic to the goal exceeds its running minimum by at least
2\,m before its first step inside the goal region. Recovery is the share of these episodes that later
enter the region.

Table~\ref{tab:supp-cap} reports the reach term and the stop term of the SR gain, together with the gain
in recovery, for one round and two rounds of \method on both benchmarks.
Two rounds of \method improve all three skills on both benchmarks. On \reverie they give a reach term
of 3.8\%, a stop term of 11.8\% and a recovery gain of 7.4\%. On NavRAG-CE they give a reach term of
0.9\%, a stop term of 10.1\% and a recovery gain of 6.9\%, so all three skills carry over to a
benchmark that \method never trains on. One round already gives a stop term of 11.6\% and a recovery
gain of 2.2\% on \reverie, and a stop term of 11.2\% and a recovery gain of 2.8\% on NavRAG-CE.

The second round lifts reaching and recovery on both benchmarks. On \reverie it reaches the goal
region in 195 more episodes than the first round, and on NavRAG-CE in 220 more.

\section{Agent sessions}
\label{sec:supp-sessions}
\begin{table}[t]
  \centering
  \tablestyle[4.0pt]
  \caption{\textbf{Planner tools} and their arguments.}
  \label{tab:supp-tools}
  \newcommand{\toolname}[1]{\texttt{\begingroup\hyphenpenalty=10000\exhyphenpenalty=10000#1\endgroup}}
  \begin{tabular}{@{\hspace{\tabcolsep}}>{\raggedright\arraybackslash}p{0.33\linewidth}>{\raggedright\arraybackslash}p{0.22\linewidth}>{\raggedright\arraybackslash}p{0.35\linewidth}@{\hspace{\tabcolsep}}}
    \toprule
    Tool & Arguments & What it does \\
    \midrule
    \toolname{navigate\_\allowbreak by\_\allowbreak instruction} & instruction text
      & The VLA drives. Returns waypoint photographs, four end views and the map. \\
    \toolname{navigate\_\allowbreak relative} & turn\_deg, distance\_m
      & Walks on a bearing, positive is left. Returns the walked distance and four views. \\
    \toolname{navigate\_\allowbreak to\_\allowbreak node} & node number or name
      & Retraces ground already covered back to a recorded node. \\
    \toolname{observe\_\allowbreak panorama} & --
      & Four labelled views at the current pose. The robot does not move. \\
    \toolname{observe\_\allowbreak forward} & --
      & Forward camera at full size. The robot does not move. \\
    \toolname{observe\_\allowbreak map} & --
      & Top-down view of the explored region and the list of recorded nodes. \\
    \toolname{annotate\_\allowbreak node} & name, caption
      & Names the current node so that the planner can return to it. \\
    \toolname{terminate\_\allowbreak episode} & here or a node
      & Ends the episode where the robot stands or at a recorded node. \\
    \bottomrule
  \end{tabular}
\end{table}

\begin{table}[t]
  \centering
  \tablestyle[3.2pt]
  \caption{\textbf{Second-round sessions by the first-round failure class.} A \emph{state} is one
  failed run, paired with three VLA seeds. Budget left and New steps are medians over states and over
  logged runs.}
  \label{tab:supp-sess}
  \begin{tabular}{@{\hspace{\tabcolsep}}lrrrr@{\hspace{\tabcolsep}}}
    \toprule
    Class   & States & Pairs & Budget left & New steps \\
    \midrule
    Never   & 177 & 531 & 452 & 66 \\
    Left    &  39 & 117 & 414 & 49 \\
    \midrule
    All     & 216 & 648 & 446 & 62 \\
    \bottomrule
  \end{tabular}
\end{table}

\begin{figure}[t]
  \centering
  \includegraphics[width=\linewidth]{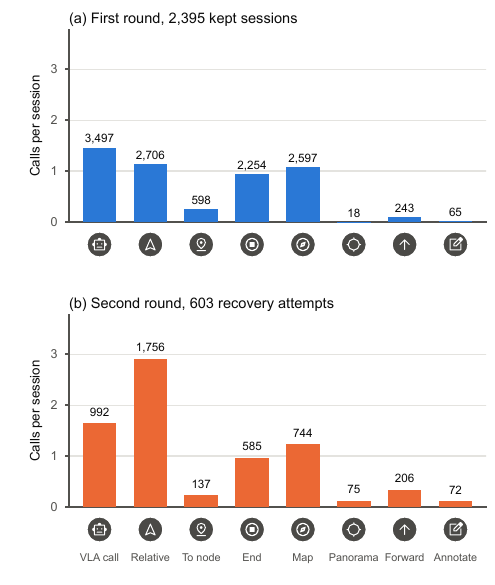}
  \caption{\textbf{Planner tool calls in each round of sessions.} \textbf{(a)} first round, over
  the \num{2395} kept sessions. \textbf{(b)} second round, over the 603 recovery attempts with a
  logged tool-call trace. Bars give calls per session and labels give the total.}
  \label{fig:supp-tools}
\end{figure}

\begin{figure*}[!t]
  \centering
  \includegraphics[width=\textwidth]{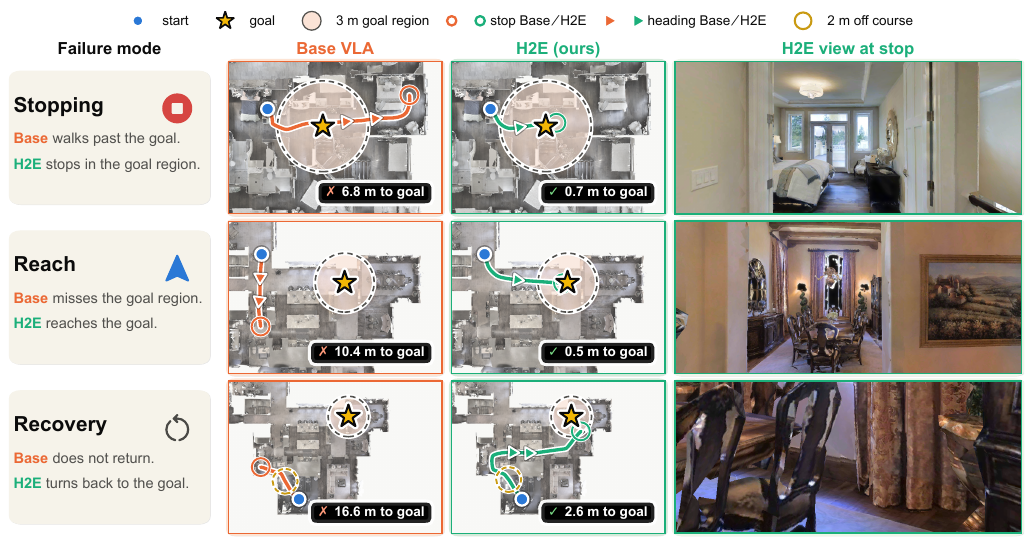}
  \caption{\textbf{Stopping, reach and recovery on NavRAG-CE val-unseen.} Each row pairs the base VLA path and
  the \method path from the same start, with the \method view at its stop.}
  \label{fig:supp-nrag-qual}
\end{figure*}

\begin{figure*}[!t]
  \centering
  \includegraphics[width=\textwidth]{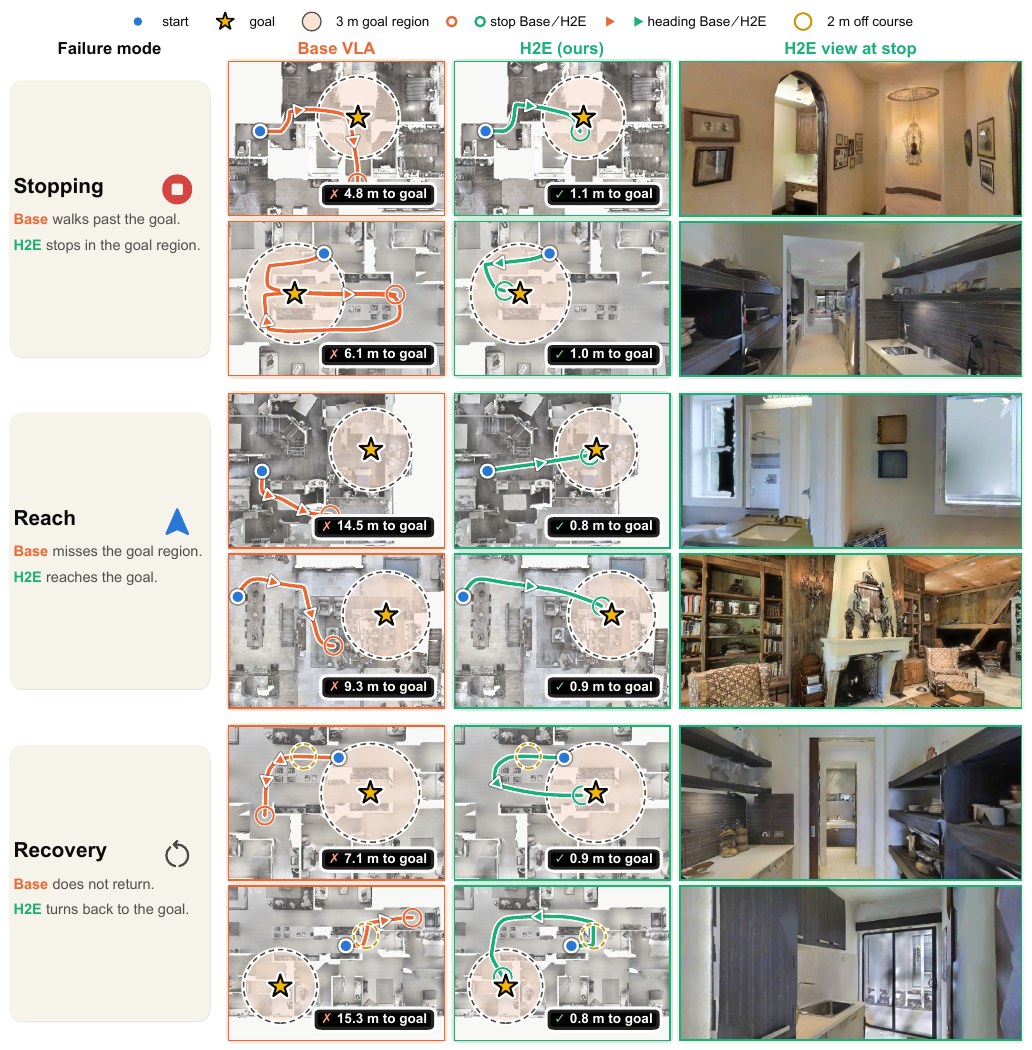}
  \caption{\textbf{Two more examples per capability on \reverie val-unseen.} Each row pairs the base VLA path
  and the \method path from the same start.}
  \label{fig:qual-more}
\end{figure*}

\paragraph{Planner tools and briefing.}
The agent of \mainref{sec:harness} talks to the environment only through the named tools of
Tab.~\ref{tab:supp-tools}. Each tool has a fixed set of arguments. The planner reads a short
description of every tool at the start of the episode.

The planner is the same LLM named in \mainref{sec:exp-setup}. Its briefing spells out how the
tools divide the work. The VLA drives. The node graph remembers. The planner decides whether the
goal has been reached. The briefing names seven numbered steps for a typical episode. Step 1
calls \emph{navigate\_by\_instruction} with $\ell$. Step 2 compares the waypoint photographs and
the four end views with the goal that the instruction describes. Step 3 looks for the room and the
object that the instruction names. Step 4 calls \emph{observe\_map} once to read every recorded node.
Step 5 goes to a node or walks the last few metres if a better match for the goal is near. Step
6 calls \emph{navigate\_by\_instruction} again with a remaining instruction when the goal has
not been seen. Step 7 calls \emph{terminate\_episode} once the robot is at the goal.

The briefing never names the goal in world coordinates and never passes the geodesic distance to
the planner. The planner decides what fits the goal only from the returned photographs, the node
list and its own memory of the episode.

\paragraph{Two rounds of sessions.}
The first round runs the agent of \mainref{sec:harness} on \num{4682} REVERIE-CE training
episodes from the original start pose, under the default budget of 500 low-level steps. We keep
the \num{2395} sessions that reach the goal. These sessions span 57 training scenes. The planner
repairs 857 of them with later tool calls, and these sessions supply 65.8\% of the labelled steps of
\mainref{tab:anatomy}. The median kept session runs 76 environment steps and walks
13.9\,m. Across the kept sessions the planner issues \num{2706} calls to \emph{navigate\_relative},
598 calls to \emph{navigate\_to\_node} and \num{2254} calls to \emph{terminate\_episode}.

The second round starts from states where the first-round VLA fails. We select 935 training
episodes, run the first-round VLA on each from its own start,
and collect every episode in which the VLA ends more than 3\,m from the goal with a valid pose and
at least 20 remaining steps. We split these states into two classes. In the \emph{never} class the
VLA never comes within 3\,m of the goal during the failed run. In the \emph{left} class the VLA
enters the 3\,m region and then leaves it. We keep at most one state per episode and class, which
gives 216 unique states, 177 of the \emph{never} class and 39 of the \emph{left} class. The states
come from 18 training scenes.

Table~\ref{tab:supp-sess} lists the two classes and the sessions collected on them. We pair each
state with three VLA seeds, 20261028, 20261029 and 20261030, which gives 648 state-seed pairs for the
agent. The agent sees the same tools and the same briefing as in the first
round. The only difference is the first prompt, which tells the planner that the robot already
walked part of the episode, states the actual remaining step budget, and asks it to call
\emph{observe\_panorama} first. The median \emph{never} state keeps 452 steps of budget, the
median \emph{left} state keeps 414 steps. In total, 435 recovery sessions reach the goal
and enter training alongside the first-round sessions.

Figure~\ref{fig:supp-tools} splits the tool calls of the two rounds by tool. The second round calls
\emph{navigate\_relative} two to three times more often per session than the first round, which
matches a mid-episode restart from a hard state. It also looks around more, with 1.7 observe
calls per attempt against 1.2 per session in the first round.

\paragraph{Collection cost.}
The agent sessions are a one-time offline cost paid before training. The first round runs
\num{4682} sessions. The agent reaches the goal in \num{2395} of them and fails in \num{2287}, and the
planner makes \num{26563} tool calls across all of them. The second round
runs the agent on 648 state-seed pairs, and its 603 attempts with a logged trace make \num{4567} tool
calls. After training, the VLA runs without the agent and uses no LLM tokens
at test time, as \mainref{tab:harness} shows.

\section{Qualitative examples}
\label{sec:supp-qual}
Figure~\ref{fig:supp-nrag-qual} shows one pair per capability on NavRAG-CE, the benchmark that \method never trains
on. \method stops inside the goal region, reaches a goal that the base VLA misses, and turns back to the
goal after going off track. Figure~\ref{fig:qual-more} shows two more pairs per capability on other \reverie
val-unseen episodes, paired as in \mainref{fig:qual}.

\section{Implementation details}
\label{sec:supp-impl}
\paragraph{Evaluation protocol.} The released Dynam3D code stops the robot after 50 waypoint decisions and scores its final position. Our
VLA takes up to 500 low-level steps and must stop on its own within the 2{,}400\,s limit of the official
evaluation script. Both protocols use the same 3\,m success radius.

\paragraph{Sessions.} The first 504 kept sessions make 577 calls to \emph{navigate\_relative} and 142
calls to \emph{navigate\_to\_node}. Of these sessions, 289 contain sub-instruction or planner-tool
steps. In 330 sessions the robot ends within 1\,m of the goal. In 121 sessions it passes within 1\,m
and ends 1 to 3\,m away; for the states after its last end label we replace the target of
Eq.~\ref{eq:rc} with a geodesic target. In the other 53 sessions the last targets follow the geodesic
to the 1\,m boundary.

\paragraph{Ablation data.} Without task-level labels, the trainable agent steps drop from \num{55504} to
\num{39013} and the end labels drop from \num{7322} to \num{4744}. The ablation without end labels keeps
the 336 end labels at the final poses of 330 sessions, and its trainable agent steps drop to \num{48518}. The
ablation with geodesic actions replaces the \num{48182} motion targets that are not end labels. Self-imitation uses the 362 successful runs of the base VLA on
the 935 training episodes used for second-round state collection.

\paragraph{Optimization.} \Mainref{tab:setup} lists the settings of \method. The adapted VLAs of
\mainref{tab:recipe} use the same learning rates and train for \num{3000} updates, or \num{48000} samples. We
keep a checkpoint every 500 updates and pick one on 30 other training episodes.

\paragraph{Data splits.} No training scene appears in val-unseen. The models of \mainref{tab:harness,tab:recipe} do
not train on the scenes of the 100 held-out episodes.

\paragraph{Proof of Proposition 1.}
Fix a history $h$ and write $\pi(h)\in\mathbb{R}^{K\times 3}$ coordinate-wise. The conditional expectation
$\mathbb{E}\big[\lVert \pi(h)-y\rVert_1\mid h\big]$ equals $\sum_j \mathbb{E}\big[\lvert\pi(h)_j-y_j\rvert\mid h\big]$.
Each term is minimized by any conditional median of $y_j$ given $h$, so every coordinate of a minimizer is
a median of the corresponding coordinate of $y$. If $P(y=\mathbf{0}\mid h)>1/2$, then $P(y_j=0\mid h)>1/2$
for every $j$, so $0$ is the unique median of each coordinate and $\mathbf{0}$ is the unique minimizer.
\hfill$\square$

\else
\twocolumn[{%
\renewcommand\twocolumn[1][]{#1}%
\maketitle
\begin{center}
  \captionsetup{type=figure}
  \includegraphics[width=\textwidth]{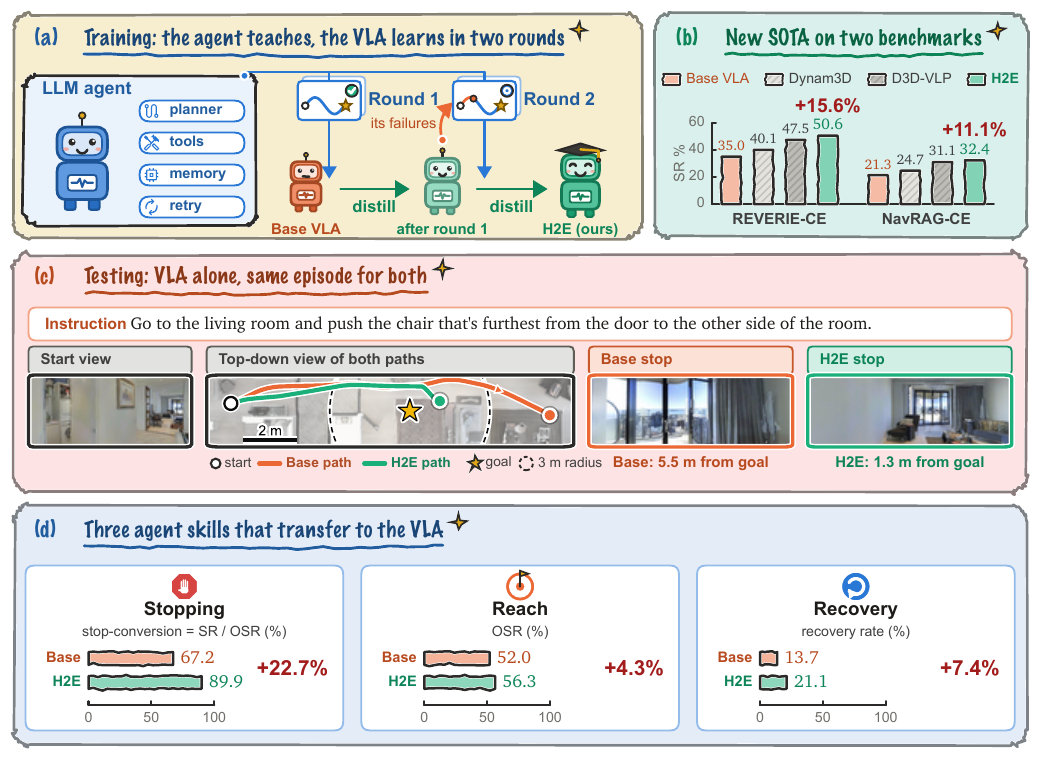}
  \vspace{-1.2em}
  \caption{\textbf{Training with the agent, testing the VLA alone.} \textbf{(a)}~\method distills the
  successful sessions of an LLM agent into the VLA it calls, in two rounds. \textbf{(b)}~Success rate (SR)
  on \reverie and NavRAG-CE val-unseen. \textbf{(c)}~At test time the VLA runs alone, and \method stops
  inside the goal region. \textbf{(d)}~Three agent skills that transfer to the VLA.}
  \label{fig:teaser}
\end{center}
\vspace{0.6em}
}]
\ifdefined\cvprsubmit\else
{\renewcommand{\thefootnote}{\fnsymbol{footnote}}\footnotetext[1]{Corresponding author.}}
\fi

\begin{abstract}
Vision-language-action (VLA) models drive the robot directly from camera views, yet a VLA alone shows three
typical failures. It walks past the goal, stops before reaching it, and rarely recovers after going off
track. Pairing a VLA with a large language model (LLM) planner fixes these failures, but queries the LLM
throughout every episode, adding token costs and latency. We introduce Harness-to-Executor (\method),
a task-level distillation method that pairs an LLM planner and a frozen VLA as a heterogeneous
training agent, keeps only sessions that reach the goal, and then trains a VLA alone on them so no LLM
is used at test time. In
these sessions, the planner gives sub-instructions and moves the robot with its own tools, detours are
common, and the planner decides when to stop. \method therefore labels every motion as a VLA action for
the original task, whoever made it. It masks the loss on steps moving away from the goal and assigns end
labels to states near it, so the VLA learns where the task ends. \method sets a new state of the art on
\reverie and NavRAG-CE. It
raises an 8B VLA's success rate by 15.6\% and 11.1\% respectively, surpassing a method trained on 10M
samples. The distilled VLA exceeds its agent on unseen episodes with zero LLM tokens and runs 3.7 times faster.
\end{abstract}

\section{Introduction}
\label{sec:intro}

Goal-directed navigation asks a robot to find a remote object that an instruction describes, such as
\emph{go to the second-floor hallway and pick up the picture above the table}~\cite{qi2020reverie}. A
vision-language-action (VLA) model can drive the robot directly from its camera views and the
instruction~\cite{navgpt3,navid,wei2025streamvln}. A VLA runs at the pace of the camera, keeps the robot in
one network, and needs no external memory.

A VLA alone, however, often fails in a goal-directed episode. Three failures are typical, and we use
\reverie~\cite{qi2020reverie,wang2025dynam3d} as our running case. First, the VLA walks past the goal: it
reaches the 3\,m success region of the goal but keeps moving and leaves it~\cite{xfly,stopwatch}. Second,
the VLA stops before reaching the goal: once the goal is out of view it issues a stop although it never
reached the success region~\cite{realm}. Third, the VLA does not recover: after it goes off track, it
rarely returns to search for the goal~\cite{budvln}.

Recent systems avoid these failures by pairing the VLA with a large language model (LLM)
planner~\cite{navgpt3,navmcp}. The planner reads the scene and keeps a map of visited places. It hands
short goals to the VLA, moves the robot itself when needed, and decides when to stop. The agent raises the
success rate of the VLA, but it calls the LLM throughout every episode. Each episode
therefore spends many LLM tokens and runs several times longer than the VLA alone.

We ask whether the VLA can learn the skills of the agent from the sessions the agent records and then
navigate alone. Current agents keep the VLA frozen~\cite{navgpt3}, and distillation from such agents has
so far compressed the planner~\cite{tinyastral,omnibrain0,guava}. We instead train the VLA.

A session of the agent is hard to use as training data for the VLA, for three reasons. The planner often
moves the robot with its own tools or hands the VLA new sub-instructions, and these produce much of the
motion in a session. The sessions contain detours that the planner later fixes, so many steps move away
from the goal. The planner, not the VLA, decides when a session ends. These mismatches mirror the three
failures of the VLA one for one, and together they shape the three rules of our method.

We present Harness-to-Executor (\method), a task-level experience distillation method that transfers the
skills of the agent harness to its executor, the VLA. Figure~\ref{fig:pipeline} gives an overview. First,
\method labels every motion in a session as a VLA action under the original task, whoever made the motion.
Second, it masks the loss on steps that move away from the goal. Third, it labels every visited state near
the goal as an end of the session. Low-rank adaptation~\cite{hu2022lora} then trains the VLA on these labels together
with standard geodesic-expert data~\cite{laskey2017dart,anderson2018r2r,krantz2020vlnce}. At test time the
VLA replaces the base VLA and runs alone.

Our contributions are as follows.
\begin{itemize}
  \item We distill an LLM navigation agent into the VLA that it calls. The VLA then navigates alone and
  needs no LLM at test time.
  \item We present \method, which turns successful agent sessions into VLA training data by three rules:
  task-level labels, progress masking and end labels. Each rule answers one way in which a session
  differs from a VLA demonstration.
  \item \method sets a new state of the art on two goal-directed navigation benchmarks in unseen scenes.
  The distilled VLA stops at the goal, keeps its reach, and learns to
  recover after going off track. It beats the success rate of the agent that taught it, with no LLM tokens
  and in far less time.
\end{itemize}

\section{Related Work}
\label{sec:related}
\begin{figure*}[t]
  \centering
  \includegraphics[width=\textwidth]{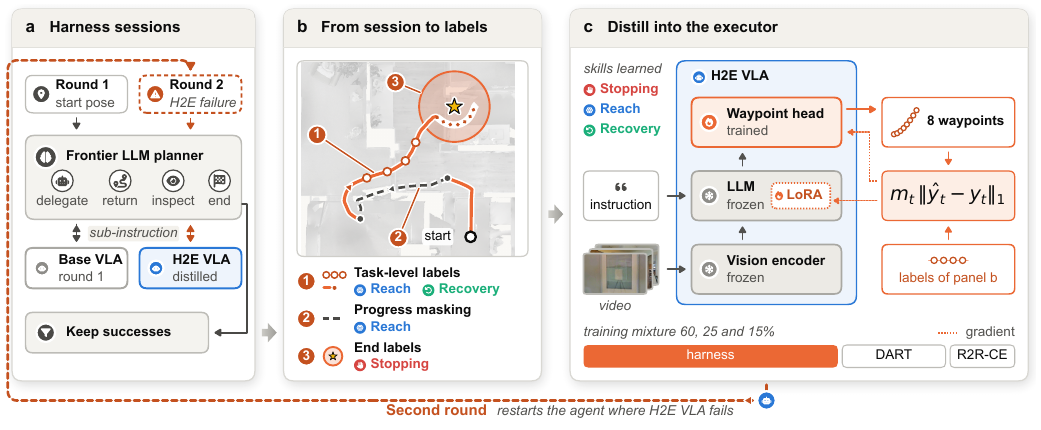}
  \caption{\textbf{Overview of \method.} \textbf{(a)}~An LLM planner calls the VLA on training scenes,
  and we keep sessions that reach the goal. The second round starts from failures of the
  first-round VLA. \textbf{(b)}~Three rules turn each session into labels for the three skills.
  \textbf{(c)}~The VLA trains LoRA adapters and its action head on a 0.60/0.25/0.15 mix
  of agent sessions, DART and R2R-CE, then runs alone at test time.}
  \label{fig:pipeline}
\end{figure*}

\paragraph{VLA models for navigation.}
Navigation in continuous environments~\cite{krantz2020vlnce} has moved from waypoint predictors to video
VLA models that map camera views and an instruction to
actions~\cite{navid,uninavid,navila,wei2025streamvln,navgpt3}. On \reverie, 3D-token models trained on
millions of samples lead the benchmark~\cite{wang2025dynam3d,d3dvlp}. Other work improves stopping with
grounding modules or with stop labels on the learner's own states~\cite{realm,stopwatch,xfly}. These
policies learn from geometric experts or offline data. Our VLA learns from the agent that calls it.

\paragraph{LLM agents and their students.}
LLM agents that call navigation models as tools lead several instruction-following
benchmarks~\cite{navgpt3,navmcp,qwenrobotnav}. Their improvement loops edit the planner's context and keep
the VLA fixed~\cite{navgpt3}. Distillation from these agents has targeted the planner. TinyAstral
replaces the planner with a 4B student~\cite{tinyastral}, Guava distills agent trajectories into a compact
agent~\cite{guava}, and OmniBrain-0 moves the planner into a smaller model~\cite{omnibrain0}. Harness-Zero
maps agent traces into the action space of another agent~\cite{harnesszero}. Other systems adapt the VLA
to the planner's interface~\cite{dualvln}, train both levels with reinforcement learning (RL)~\cite{match},
or relabel VLA segments with planner sub-instructions~\cite{decoupleplan}. We keep the whole agent as the
teacher, train only the VLA, and remove the LLM at test time. We keep the task instruction and credit the
VLA with every motion, whereas HSL and Phi-Nav rewrite the instruction and Harness-Zero maps the
trace to another agent~\cite{hsl,phinav,harnesszero}.

\paragraph{Capability transfer from a privileged teacher.}
OEC switches to the expert inside student rollouts~\cite{oec}, and SAGE distills a vision-language teacher
into an RL policy~\cite{sage}. A strong privileged teacher can also teach a memoryless student
poorly~\cite{leap}. These studies ask whether a teacher's skill transfers at all. We split navigation into
three skills, stopping, reaching and recovering, and show that a VLA takes over all three from the LLM agent
that calls it, with no module added at test time.

\paragraph{Relabelling demonstrations.}
HSL relabels agent trajectories with the goal they reached~\cite{hsl}. Phi-Nav writes new instructions for
on-policy vision-and-language navigation (VLN) paths~\cite{phinav}, and BudVLN shows that recovery labels can conflict with the original
instruction~\cite{budvln}. In object-goal navigation, human and exploration demonstrations transfer better
than shortest paths~\cite{habitatweb,pirlnav}, and self-generated paths help goal-directed
VLN~\cite{sid,srdf}. HSL and Phi-Nav change the instruction to fit the path. \method keeps the task
instruction, credits the VLA with every motion of the session, and masks steps by their progress toward the
goal.

\section{Method}
\label{sec:method}

\method distills the successful sessions of an LLM agent into the VLA it calls, and the VLA then runs
alone. We run the agent on training scenes and keep the sessions that reach the goal. Three rules turn
them into samples for a LoRA update of the VLA. An optional second round restarts the agent where the
trained VLA fails. Figure~\ref{fig:pipeline} and Algorithm~\ref{alg:h2e} summarize \method.

\paragraph{Setup.}
An episode in \reverie gives a natural-language instruction $\ell$ that refers to a remote object, a start
pose and a goal $g$ near that object. At step $t$ the robot stands at pose $p_t$, receives four RGB views
$o_t$ and takes an action. Let $d(p)$ be the geodesic distance from $p$ to $g$. An episode succeeds if the
robot stops with $d(p_T)\le 3$\,m. Each episode allows at most 500 steps.

\paragraph{VLA.}
The VLA $\pi$ is the 8B NavGPT video VLA~\cite{navgpt3}. It reads its last 16 observations and the
instruction, $h_t=(o_{t-15:t},\ell)$, and predicts $K = 8$ waypoints $\hat{y}_t=\pi(h_t)\in\mathbb{R}^{K\times
3}$. Each waypoint holds a planar offset and a heading in the robot frame. The robot executes the first
waypoint and then queries the VLA again. The VLA has no stop head. It stops when two outputs in a row are
all zero. Moving and stopping thus come from one regression output, so the labels near the goal teach both.

\paragraph{Agent.}
The agent $\mathcal{H}$ pairs an LLM planner with the VLA as a tool~\cite{navgpt3}. The planner
can send an instruction or a sub-instruction to the VLA. It can also move the robot itself or walk
back to a place in its place graph. It can look at a panorama or the map, and it can end the episode.
The planner decides with $h_t$ and with extra information $z_t$ that the VLA never sees, namely the
place graph and the whole episode history.

\subsection{Agent sessions as task-level experience}
\label{sec:harness}
A session mixes VLA actions, planner-tool motions and new sub-instructions, and we credit every motion to
the VLA under the original instruction $\ell$.
We run $\mathcal{H}$ on training episodes from their original start pose. Each session is a record
\begin{equation}
  \tau=\{(p_t,\,o_t,\,\ell_t,\,c_t)\}_{t=0}^{T}.
  \label{eq:session}
\end{equation}
Here $c_t\in\{\cexec,\ctool\}$ says whether the VLA or a planner tool moved the robot from $p_t$ to
$p_{t+1}$, and $\ell_t$ is the instruction the VLA received at that step. We split planner tool motions
into environment steps, render the views at the visited poses, and keep the \num{2395} sessions that
reach the goal, from 57 training scenes.

Table~\ref{tab:anatomy} pools the \num{332405} labelled steps and \num{30181} end labels of these sessions
across four actors. The VLA on $\ell$ takes only 53.9\% of the steps, so Eq.~\ref{eq:rc} draws the other
\textbf{46.1\%} of the steps from the other actors. Rescued sessions supply \textbf{65.8\%} of
all labelled steps, so most data comes from calls the agent repaired. Planner tools alone add
\textbf{31.8\%} of the end labels at poses where the VLA by itself never stops.
\begin{table}[t]
  \centering
  \tablestyle[3.0pt]
  \caption{\textbf{Who moves and stops the robot}, pooled over all \num{2395} kept sessions. Tail steps
  end a VLA call, and masked steps leave the loss by Eq.~\ref{eq:mask}.}
  \label{tab:anatomy}
  \begin{tabular}{@{}lrrr>{\hspace{2pt}}r>{\hspace{4pt}}r@{}}
    \toprule
    & \multicolumn{2}{c}{VLA} & \multicolumn{1}{c}{Planner} & & \\
    \cmidrule(lr){2-3}\cmidrule(lr){4-4}
    & \multicolumn{1}{c}{on $\ell$} & \multicolumn{1}{c}{on $\ell_t\neq\ell$} & \multicolumn{1}{c}{Tools}
    & \multicolumn{1}{c}{Tail} & \multicolumn{1}{c}{All} \\
    \midrule
    Step share (\%)           & 53.9 & 20.5 & 24.6 & 1.1 & 100 \\
    \subrow from rescued (\%) & 45.4 & 96.4 & 85.5 & 52.8 & \textbf{65.8} \\
    \subrow masked (\%)       & 27.3 & 24.1 & 14.7 & 5.1 & 23.3 \\
    End-label share (\%)      & 50.8 & 12.8 & \textbf{31.8} & 4.5 & 100 \\
    \bottomrule
  \end{tabular}
\end{table}

\subsection{From sessions to a standalone VLA}
\label{sec:labels}
Three rules turn each session into training samples $x_t=(o_{\le t},\ell,y_t,m_t)$ with a history, an
instruction, a target and a mask.

\paragraph{Task-level labels.}
Let $\Phi_K$ express the next $K$ poses in the robot frame at $p_t$ with the waypoint spacing of the VLA.
\method labels every step as
\begin{equation}
  \begin{aligned}
    \mathcal{R}(\tau) &= \big\{(o_{\le t},\,\ell,\,y_t)\big\}_{t<T},\\
    y_t &= \Phi_K(p_t;\,p_{t+1:t+K}).
  \end{aligned}
  \label{eq:rc}
\end{equation}
The label depends neither on $c_t$ nor on $\ell_t$, so every motion of the session becomes a VLA action
under the test-time instruction. The target is simply where the session went next.

\paragraph{Progress masking.}
Sessions contain detours that the agent later fixes. Step $t$ enters the loss, $m_t=1$, if one of
$p_t,\dots,p_{t+K}$ lies near the goal by Eq.~\ref{eq:term} or if
\begin{equation}
  d(p_{t+K}) \le d(p_t) + \delta,
  \label{eq:mask}
\end{equation}
with $\delta = 0.25$\,m. A turn in place keeps $d$, so it stays in the loss unless it starts a detour.
Masked steps stay in the history
$o_{\le t}$, and only their targets leave the loss. The distance $d$ is used only to build labels, and the
VLA never receives it.

\paragraph{End labels.}
The planner, not the VLA, decides when a session ends. End labels carry this decision across, marking the
states near the goal as the end of the task so the VLA learns where the task ends. One rule assigns an
end label for every data source, including the geodesic expert,
\begin{equation}
  y_t=\mathbf{0}\quad \text{if } d(p_t)\le \rarr,
  \label{eq:term}
\end{equation}
with $\rarr = 1$\,m. A radius tighter than the 3\,m success region places the end labels well inside it.
The rule also adds end labels at states that the VLA walked through without stopping. This happens when
the first VLA call passes within 1\,m of the goal and then leaves the goal region.

\begin{algorithm}[t]
  \caption{Task-level experience distillation with \method}
  \label{alg:h2e}
  \small
  \begin{algorithmic}[1]
    \Require episodes $\mathcal{E}$, agent $\mathcal{H}$, VLA $\pi_{\theta_0}$, data
    $\mathcal{D}_{\mathrm{DART}},\mathcal{D}_{\mathrm{R2R}}$
    \State $\mathcal{S}\gets\{\tau\sim\mathcal{H}(e) : e\in\mathcal{E},\ \tau \text{ succeeds}\}$
      \Comment{Eq.~\ref{eq:session}}
    \State $\mathcal{D}_{\mathcal{H}}\gets\emptyset$
    \For{$\tau\in\mathcal{S}$ \textbf{and} $(o_{\le t},\ell,y_t)\in\mathcal{R}(\tau)$}
      \Comment{Eq.~\ref{eq:rc}}
      \State \textbf{if} $d(p_t)\le \rarr$ \textbf{then} $y_t\gets\mathbf{0}$ \Comment{Eq.~\ref{eq:term}}
      \State $m_t\gets$ progress mask \Comment{Eq.~\ref{eq:mask}}
      \State $\mathcal{D}_{\mathcal{H}}\gets\mathcal{D}_{\mathcal{H}}\cup\{x_t\}$
    \EndFor
    \State $\mathcal{D}\gets$ mix of $\mathcal{D}_{\mathcal{H}},\mathcal{D}_{\mathrm{DART}},\mathcal{D}_{\mathrm{R2R}}$
      at 0.60\,:\,0.25\,:\,0.15
    \State $\theta\gets$ LoRA update of $\theta_0$ minimizing Eq.~\ref{eq:loss} on $\mathcal{D}$
    \State \Return $\pi_\theta$ \Comment{runs alone at test time}
  \end{algorithmic}
\end{algorithm}

\begin{table*}[t]
  \centering
  \tablestyle[2.2pt]
  \sbox{\tabmainbox}{%
  \begin{tabular}{l>{\hspace{2pt}}ll>{\hspace{3pt}}cccc>{\hspace{3pt}}cccc}
    \toprule
     &  &  & \multicolumn{4}{c}{REVERIE-CE val-unseen} & \multicolumn{4}{c}{NavRAG-CE val-unseen} \\
    \cmidrule(l{3pt}r{3pt}){4-7}\cmidrule(l{3pt}r{3pt}){8-11}
    Method & Observation & Training data & SR$\uparrow$ & OSR$\uparrow$ & SPL$\uparrow$ & NE$\downarrow$ & SR$\uparrow$ & OSR$\uparrow$ & SPL$\uparrow$ & NE$\downarrow$ \\
    \midrule
    \multicolumn{11}{l}{\gray{\textit{Published, monocular setting}}} \\
    \gray{InstructNav~\cite{instructnav}$^{\ast}$} & \gray{RGB-D} & \gray{zero-shot}       & \gray{25.2} & \gray{31.5} & \gray{19.1} & \gray{7.44} & \gray{17.4} & \gray{24.1} & \gray{10.9} & \gray{9.83} \\
    \gray{NaVid~\cite{navid}$^{\dagger}$}           & \gray{RGB}   & \gray{Dynam3D mixture} & \gray{26.6} & \gray{36.3} & \gray{20.8} & \gray{6.74} & \gray{19.4} & \gray{29.6} & \gray{13.9} & \gray{9.35} \\
    \gray{g3D-LF~\cite{g3dlf}$^{\dagger}$}          & \gray{RGB-D} & \gray{Dynam3D mixture} & \gray{34.4} & \gray{41.6} & \gray{23.8} & \gray{6.50} & \gray{21.4} & \gray{31.8} & \gray{13.5} & \gray{8.85} \\
    \gray{Dynam3D~\cite{wang2025dynam3d}}           & \gray{RGB-D} & \gray{4M samples}      & \gray{40.1} & \gray{48.9} & \gray{28.5} & \gray{6.22} & \gray{24.7} & \gray{38.4} & \gray{18.8} & \gray{8.12} \\
    \gray{D3D-VLP~\cite{d3dvlp}}                    & \gray{RGB-D} & \gray{10M samples}     & \gray{47.5} & \gray{56.9} & \gray{34.7} & \gray{5.36} & \gray{31.1} & \gray{45.3} & \gray{23.9} & \gray{7.57} \\
    \midrule
    \blockgap
    \multicolumn{11}{l}{\textit{8B NavGPT VLA~\cite{navgpt3}, running alone}} \\
    Base VLA                      & 4-view RGB & --                                  & 35.0 & 52.0 & 26.5 & 8.53 & 21.3 & 60.8 & 14.1 & 9.41 \\
    + Geodesic adaptation         & 4-view RGB & DART~\cite{laskey2017dart} + R2R    & 34.9 & 38.5 & 33.2 & 8.93 & --   & --   & --   & --   \\
    \oursrow
    + \method, one round          & 4-view RGB & \num{2395} sessions + DART + R2R    & 44.9 & 50.1 & 38.4 & 6.90 & 31.8 & 59.4 & 23.8 & 7.94 \\
    \oursrow
    + \method, two rounds         & 4-view RGB & \;\;+435 recovery sessions          & \best{50.6} & 56.3 & \best{40.8} & 6.26 & \best{32.4} & \best{62.6} & 20.6 & 7.80 \\
    \bottomrule
  \end{tabular}}%
  \captionsetup{width=\wd\tabmainbox}
  \caption{\textbf{Comparison on two goal-directed unseen splits.} \method sets a new state of the art on
  \reverie and NavRAG-CE and surpasses D3D-VLP, which uses 10M training samples. Published rows are copied
  from Dynam3D~\cite{wang2025dynam3d} (Tab.~2) and D3D-VLP~\cite{d3dvlp} (Tab.~2). $^{\ast}$Zero-shot.
  $^{\dagger}$Retrained on the Dynam3D data.}
  \label{tab:main}
  \usebox{\tabmainbox}

  \vspace{4pt}
  \parbox{\wd\tabmainbox}{\tablenote REVERIE-CE uses all \num{3189} and NavRAG-CE all \num{6932} val-unseen episodes. Geodesic adaptation
  was not run on NavRAG-CE.}
\end{table*}

\paragraph{Training objective.}
\label{sec:training}
The VLA learns from three sources. $\mathcal{D}_{\mathcal{H}}$ holds the labelled agent steps.
$\mathcal{D}_{\mathrm{DART}}$ holds geodesic expert steps with the noise of Disturbances for Augmenting
Robot Trajectories (DART)~\cite{laskey2017dart}. $\mathcal{D}_{\mathrm{R2R}}$ holds path-following steps of
Room-to-Room in continuous environments (R2R-CE)~\cite{anderson2018r2r,krantz2020vlnce}. Both come from
the training scenes and use Eq.~\ref{eq:term}. The VLA minimizes the masked $L_1$ loss over the mixture
\begin{equation}
  \mathcal{L}(\theta) = \sum_{s} w_s\,\mathbb{E}_{x_t\sim\mathcal{D}_s}
  \big[\, m_t\,\lVert \pi_\theta(o_{\le t},\ell) - y_t \rVert_1 \big].
  \label{eq:loss}
\end{equation}
Here $s$ runs over the three sources with weights $w_{\mathcal{H}}=0.60$, $w_{\mathrm{DART}}=0.25$ and
$w_{\mathrm{R2R}}=0.15$.

We fine-tune the VLA with LoRA~\cite{hu2022lora}. Each adapted weight matrix $W_0\in\mathbb{R}^{d\times k}$
of the language model becomes
\begin{equation}
  W = W_0 + \frac{\alpha}{r}\,BA,
  \label{eq:lora}
\end{equation}
where $B\in\mathbb{R}^{d\times r}$ and $A\in\mathbb{R}^{r\times k}$ are trained, with rank $r=16$ and
$\alpha=32$. LoRA adapts the attention and feed-forward layers. The action head is trained in
full, and the vision encoder stays frozen. The parameters $\theta$ thus hold $A$, $B$ and the head.

\paragraph{Why end labels teach stopping.}
The VLA stops only when its output is zero. The $L_1$ loss can drive the output exactly to zero near the
goal, by a classical property of the $L_1$ Bayes optimum.

\smallskip\noindent\textbf{Proposition 1.}\label{prop:median} \emph{Fix a history $h$ and let $y$ be the
target of a training sample with $m=1$. Every minimizer of $\mathbb{E}\big[\lVert \pi(h)-y\rVert_1 \mid
h\big]$ is a median of $y$ given $h$ in each coordinate. If $P(y=\mathbf{0}\mid h)>1/2$, the minimizer is
unique and equals $\mathbf{0}$.}

\smallskip\noindent
A sum of absolute errors is minimized by the conditional median in each coordinate, and a majority of
zero targets forces the median to zero. \Suppname gives the one-line proof. The end-label rule of
Eq.~\ref{eq:term} makes end labels the majority within $\rarr$ of the goal, so the $L_1$ minimizer is pulled
toward zero. A squared loss would return the mean, which stays non-zero while some samples still move.

\subsection{Second round and deployment}
\label{sec:round2}
The first round sees only states that the agent visits. To cover states where the distilled VLA itself
fails, we run it on training episodes and pick the ones where it ends far from the goal. From each we
start a recovery session of the agent. The three rules label the new sessions, and we continue training
on them with weights 0.30 for the recovery sessions, 0.30 for the first-round sessions, 0.25 for DART and
0.15 for R2R-CE. These failing states lie in training scenes outside val-unseen.

\paragraph{Deployment.}
\label{sec:deploy}
At test time the LoRA VLA replaces the base VLA and runs alone. By Eq.~\ref{eq:lora}, $BA$ can be merged into
$W_0$, so the distilled VLA has the size and step cost of the base VLA. For an episode of $T$ steps, the
time of the agent and of the distilled VLA is
\begin{equation}
  \begin{aligned}
    t_{\mathcal{H}} &= T\,c_{\pi} + \sum_{k=1}^{K} c_{\mathrm{LLM}}(n_k),\\
    t_{\method} &= T\,c_{\pi}.
  \end{aligned}
  \label{eq:cost}
\end{equation}
Here $c_\pi$ is the time of one VLA step, $K$ is the number of planner calls, and $c_{\mathrm{LLM}}(n_k)$
is the time of a call with $n_k$ tokens. The agent also pays for $\sum_k n_k$ LLM tokens per episode,
while \method pays for none. \method can also shorten $T$ because it stops at the goal.
Section~\ref{sec:exp-harness} measures these terms.

\section{Experiments}
\label{sec:exp}

\subsection{Setup}
\label{sec:exp-setup}
\paragraph{Benchmark.}
We evaluate on all \num{3189} episodes of the official \reverie~\cite{qi2020reverie,wang2025dynam3d}
val-unseen split and on all \num{6932} episodes of the NavRAG-CE~\cite{yang2024navrag} val-unseen
split. Both run in Habitat~\cite{savva2019habitat} on Matterport3D~\cite{chang2017mp3d} scenes
that no training data touches. For the comparison with the LLM agent we use 100 held-out episodes
from 26 further training scenes.
Table~\ref{tab:setup} lists the data.
\begin{table}[t]
  \centering
  \tablestyle[6pt]
  \caption{\textbf{Data and training settings of \method.}}
  \label{tab:setup}
  \begin{tabular}{@{\hspace{\tabcolsep}}ll@{\hspace{\tabcolsep}}}
    \toprule
    \multicolumn{2}{@{\hspace{\tabcolsep}}l}{\textit{Data}} \\
    \subrow Agent sessions & \num{2395} from 57 scenes \\
    \subrow Labelled agent steps & \num{332405} \\
    \subrow Held-out episodes & 100 from 26 scenes \\
    \subrow Val-unseen episodes & \num{3189} from 10 scenes \\
    \midrule
    \multicolumn{2}{@{\hspace{\tabcolsep}}l}{\textit{Model and labels}} \\
    \subrow LoRA rank $r$ and scale $\alpha$ & 16 and 32 \\
    \subrow Trained parts & LoRA and action head \\
    \subrow History frames and waypoints $K$ & 16 and 8 \\
    \subrow Mixture weights of Eq.~\ref{eq:loss} & 0.60, 0.25 and 0.15 \\
    \subrow Margin $\delta$ and end-label radius $\rarr$ & 0.25\,m and 1\,m \\
    \midrule
    \multicolumn{2}{@{\hspace{\tabcolsep}}l}{\textit{Optimization}} \\
    \subrow Samples per update & 16 \\
    \subrow Learning rate of LoRA and head & $10^{-4}$ and $2\times10^{-5}$ \\
    \subrow Warm-up updates and schedule & 540, then cosine \\
    \subrow Gradient clipping & norm 1 \\
    \subrow Updates and training time & \num{9000} and about 28\,h \\
    \bottomrule
  \end{tabular}
\end{table}

\paragraph{Metrics.}
We report SR, oracle success rate (OSR), success weighted by path length (SPL) and
navigation error (NE)~\cite{anderson2018eval}. For episode $i$ of $N$, let $S_i=\ind[d(p^i_T)\le 3]$ show
success and $R_i=\ind[\min_t d(p^i_t)\le 3]$ show that the robot reached the 3\,m goal region at some step.
Then
\begin{equation}
  \begin{aligned}
    \mathrm{SR} &= \frac{1}{N}\sum_i S_i,\qquad \mathrm{OSR} = \frac{1}{N}\sum_i R_i,\\
    \mathrm{SPL} &= \frac{1}{N}\sum_i S_i\,\frac{l_i}{\max(l_i,\,q_i)},\\
    \mathrm{NE} &= \frac{1}{N}\sum_i d(p^i_T),
  \end{aligned}
  \label{eq:metrics}
\end{equation}
where $l_i$ is the shortest path length and $q_i$ is the length of the path the robot took. An episode
ends when the VLA stops, after 500 steps, or at the 2,400\,s limit of the official evaluation script.

\paragraph{Baselines.}
All variants use the same 8B NavGPT VLA. The \emph{base VLA} is the released model. \emph{Geodesic
adaptation} trains the VLA on geodesic expert labels with DART noise~\cite{wang2025dynam3d,wei2025streamvln}
and on R2R-CE path following~\cite{anderson2018r2r,krantz2020vlnce}. It uses the same end-label rule and LoRA settings as \method. \method adds the agent sessions to the same DART
and R2R-CE data.

\paragraph{Implementation details.}
The planner of the agent is gpt-6-astra. Table~\ref{tab:setup} also lists the training settings of \method.
Training runs on four NVIDIA L40 GPUs with data parallelism, and we evaluate the final checkpoint.
\Suppname gives the evaluation protocol and further details.

\subsection{Main results}
\label{sec:exp-main}
\begin{table}[t]
  \centering
  \tablestyle[4.4pt]
  \caption{\textbf{Same training settings} on \reverie val-unseen. \method adds 504 agent sessions to the
  data of geodesic adaptation.}
  \label{tab:recipe}
  \begin{tabular}{@{\hspace{\tabcolsep}}lccccc@{\hspace{\tabcolsep}}}
    \toprule
    Method & SR$\uparrow$ & OSR$\uparrow$ & SPL$\uparrow$ & NE$\downarrow$ & $\Delta$SR \\
    \midrule
    \subrow Base VLA           & 35.0 & 52.0 & 26.5 & 8.53 & -- \\
    \subrow Geodesic adapt.    & 34.9 & 38.5 & 33.2 & 8.93 & $0.0$ \\
    \oursrow
    \subrow \method (ours)     & \best{43.2} & 46.9 & \best{38.2} & \best{7.01} & \best{$+8.2$} \\
    \bottomrule
  \end{tabular}
\end{table}

\begin{table}[t]
  \centering
  \tablestyle[6pt]
  \caption{\textbf{LoRA rank.} SR on 861 val-unseen episodes from three scenes.}
  \label{tab:rank}
  \begin{tabular}{@{}cccc@{}}
    \toprule
    Base VLA & Geodesic adapt. & \method, rank 16 & \method, rank 64 \\
    \midrule
    30.3 & 29.5 & \best{45.6} & 33.2 \\
    \bottomrule
  \end{tabular}
\end{table}

Table~\ref{tab:main} compares \method with the base VLA, geodesic adaptation and published monocular methods.
On \reverie, one round of \num{2395} agent sessions raises the SR of the VLA from 35.0\% to 44.9\%, and
geodesic adaptation gives no gain. A second round starts the agent from 216 failures of the first-round
VLA and makes 648 recovery attempts, of which 435 reach the goal and enter training as recovery sessions.
It lifts SR to 50.6\% and OSR to 56.3\%, a new state of the art that surpasses
D3D-VLP~\cite{d3dvlp}, which uses depth and 10M training samples. The gain over the base VLA is 15.6\%
and holds across scenes. The second round alone adds 5.7\% to the first.

We never collect agent sessions on NavRAG-CE~\cite{yang2024navrag}, yet \method raises SR on it
by 11.1\%, to 32.4\%, on all \num{6932} val-unseen episodes. This is also a new state of the art, above
the 31.1\% of D3D-VLP. The skills that \method distills thus carry over to a goal-directed benchmark with
other instructions in unseen scenes. On R2R-CE~\cite{anderson2018r2r,krantz2020vlnce}, whose step-by-step
directions differ from a goal, the one-round VLA keeps a high success rate of 74.1\%.

Table~\ref{tab:recipe} trains \method and geodesic adaptation with the same training settings, scenes and
number of updates, on the 504-session subset that matches the geodesic recipe. With only 504 sessions,
\method already reaches 43.2\% SR, while geodesic adaptation stays at 34.9\%. Table~\ref{tab:rank} compares two
LoRA ranks on 861 val-unseen episodes, and rank 16 is enough.

Figure~\ref{fig:analysis} shows where the gain comes from. Two rounds of \method raise SR in 9 of the 10
val-unseen scenes and in every third of the episodes by start-to-goal distance, while geodesic adaptation
helps in some scenes and hurts others. On the long third, where geodesic adaptation falls below the base
VLA, \method still leads the base VLA by 15.6\%.

\begin{figure}[t]
  \centering
  \includegraphics[width=\linewidth]{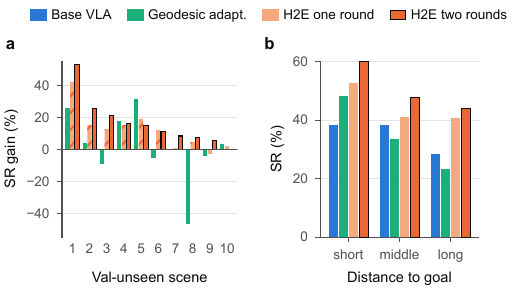}
  \caption{\textbf{Gain by scene and distance} on val-unseen. \textbf{(a)}~SR gain over the base VLA per
  scene. \textbf{(b)}~SR on thirds of the episodes by start-to-goal distance.}
  \label{fig:analysis}
\end{figure}

\subsection{Comparison with the LLM agent}
\label{sec:exp-harness}
\begin{table}[t]
  \centering
  \tablestyle[4.4pt]
  \caption{\textbf{Comparison with the LLM agent} on 100 held-out episodes from 26 unseen scenes. \method is distilled from 63
  sessions. Time and tokens are medians per episode.}
  \label{tab:harness}
  \begin{tabular}{@{\hspace{\tabcolsep}}lccccc@{\hspace{\tabcolsep}}}
    \toprule
    Method & SR$\uparrow$ & OSR$\uparrow$ & NE$\downarrow$ & Time$\downarrow$ & Tokens$\downarrow$ \\
    \midrule
    \multicolumn{6}{@{\hspace{\tabcolsep}}l}{\textit{LLM agent with the base VLA}} \\
    \subrow Agent         & 53 & \best{70} & \best{5.50} & 423\,s & 70k \\
    \midrule
    \multicolumn{6}{@{\hspace{\tabcolsep}}l}{\textit{VLA alone}} \\
    \subrow Base VLA      & 38 & 59 & 8.60 & 196\,s & 0 \\
    \subrow Geodesic adapt.  & 37 & 50 & 9.29 & 194\,s & 0 \\
    \oursrow
    \subrow \method (ours)       & \best{57} & 64 & 5.96 & \best{114\,s} & \best{0} \\
    \bottomrule
  \end{tabular}
\end{table}

Table~\ref{tab:harness} compares a VLA distilled from 63 sessions with the agent that taught it. The agent
lifts the base VLA from 38\% to 53\% SR, but it uses a median of 70k LLM tokens and 423\,s per episode.
The VLA distilled with \method reaches 57\% SR alone, turning 57 of 64 goal-region visits into success
against the agent's 53 of 70. The agent reaches the goal region more often, but the distilled VLA wins by
stopping inside it, uses no LLM tokens, and finishes in a median of 114\,s. It runs 3.7 times faster than
the agent and even faster than the base VLA. On val-unseen it needs a median of 55 steps $T$ in
Eq.~\ref{eq:cost}, against 66 for the base VLA. The distilled VLA thus keeps the skill of the agent without
its cost.

\subsection{Learnability analysis of agent skills}
\label{sec:exp-mech}
\begin{figure}[t]
  \centering
  \includegraphics[width=\linewidth]{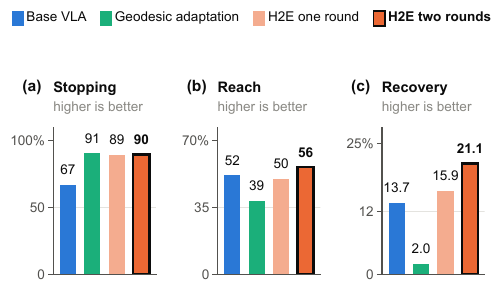}
  \caption{\textbf{Three skills on \reverie val-unseen.} Stopping is SR/OSR, reach is OSR, and recovery
  is the share of off-track episodes that return to the goal region.}
  \label{fig:mech}
\end{figure}
\begin{figure*}[t]
  \centering
  \includegraphics[width=\linewidth]{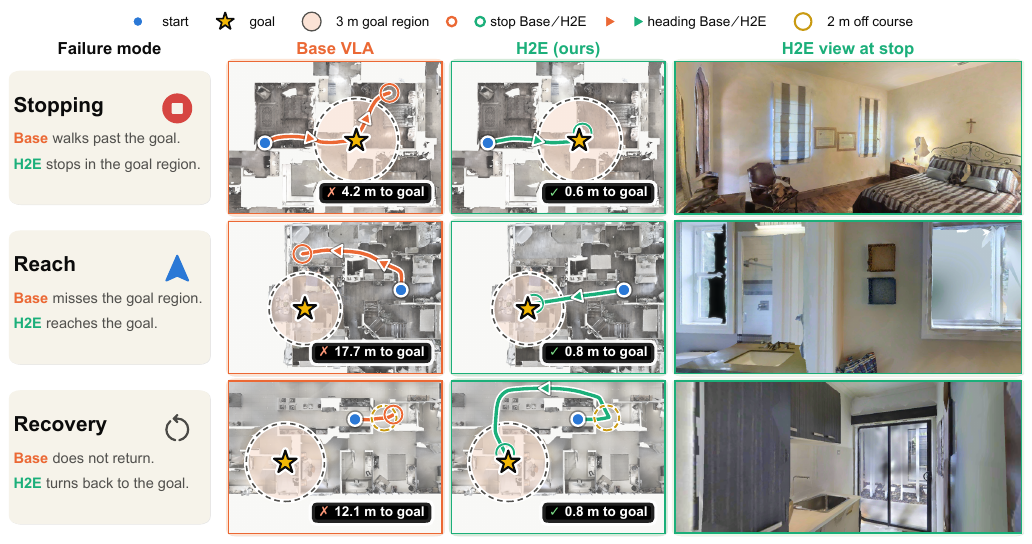}
  \caption{\textbf{Three capabilities \method transfers from the agent.} Rows correspond to
  stopping, reach and recovery. Each row pairs the base VLA path, the \method path and the view at
  \method's stop, with the final distance to the goal.}
  \label{fig:qual}
\end{figure*}
We split each outcome into reaching the goal region and stopping inside it, with $S_i$ and $R_i$ from
Eq.~\ref{eq:metrics}. For a candidate $c$ and a baseline $b$ we measure
\begin{align}
  C^c &= \frac{\sum_i S^c_i}{\sum_i R^c_i},
  \label{eq:dec-reach}\\
  N(c,b) &= \sum_i R^c_i\,(1-R^b_i),
  \label{eq:dec-n}\\
  \Delta_{\mathrm{stop}} &= \frac{1}{|A|}\sum_{i\in A}\big(S^c_i-S^b_i\big).
  \label{eq:dec-stop}
\end{align}
$C$ is the share of goal-region visits that end in success. $N$ counts the episodes that $c$ reaches and
$b$ does not. $A$ is the set of episodes that both reach, so $\Delta_{\mathrm{stop}}$ compares stopping on
the same episodes. Since SR equals OSR times $C$, the gain in SR splits into a reach term and a stop term,
\begin{equation}
  \begin{aligned}
    \mathrm{SR}^c-\mathrm{SR}^b
    &= \underbrace{(\mathrm{OSR}^c-\mathrm{OSR}^b)\,C^c}_{\text{reach term}}\\
    &\quad + \underbrace{\mathrm{OSR}^b\,(C^c-C^b)}_{\text{stop term}}.
  \end{aligned}
  \label{eq:decomp}
\end{equation}
Table~\ref{tab:decomp} gives both terms for one and two rounds of \method, plus a third skill. The
\emph{recovery rate} is the share of off-track episodes, which stray 2\,m or more before any goal visit,
that later return to the goal region. The three terms match the three capabilities named in Sec.~\ref{sec:intro}.
\begin{table}[t]
  \centering
  \tablestyle[3.6pt]
  \caption{\textbf{Three skills on REVERIE-CE val-unseen}, each as a gain over the base VLA.
  \emph{Reach} and \emph{Stop} come from Eq.~\ref{eq:decomp}. \emph{Recovery} is the share of off-track
  episodes that later return to the goal region.}
  \label{tab:decomp}
  \begin{tabular}{@{\hspace{\tabcolsep}}lcccc@{\hspace{\tabcolsep}}}
    \toprule
    Gain over the base VLA & $\Delta$SR & Reach & Stop & Recovery \\
    \midrule
    \subrow Geodesic adaptation & $0.0$  & $-12.2$ & \best{$+12.2$} & $-11.7$ \\
    \subrow \method, one round  & $+9.9$ & $-1.7$  & $+11.6$ & $+2.2$  \\
    \oursrow
    \subrow \method, two rounds & \best{$+15.6$} & \best{$+3.8$} & $+11.8$ & \best{$+7.4$} \\
    \bottomrule
  \end{tabular}
\end{table}

\paragraph{\method learns to stop.}
The base VLA ends outside the goal region in 544 of its 1\,659 reaches, converting only 67\% into
success. One round of \method raises this share to 89\% and two rounds keep it at 90\%, as
Figs.~\ref{fig:mech} and~\ref{fig:qual} show. Geodesic adaptation shares the end-label rule and stops just as well on
episodes that both reach, as Proposition~1 predicts.

\paragraph{\method keeps reaching goals and finds new ones.}
Geodesic adaptation reaches only 38.5\% of goals against 52.0\% for the base VLA. One round of \method
keeps the reach at 50.1\% and two rounds raise it to 56.3\%, with 555 new goals that the base VLA never
finds and 497 of them in success.

\paragraph{Only two rounds of \method teach recovery.}
After going off track the base VLA returns to the goal region 13.7\% of the time. Geodesic adaptation
drops this rate to 2.0\% and one round of \method keeps it at 15.9\%. Two rounds raise recovery to
21.1\%, so the recovery sessions transfer a third skill that earlier variants barely raised.

\subsection{Ablation study}
\label{sec:exp-abl}
\begin{table}[t]
  \centering
  \tablestyle[4.6pt]
  \caption{\textbf{Ablation study} on the 100 held-out episodes from 26 unseen scenes. Each row changes only the agent data. $C$
  is SR divided by OSR. EL is on for every agent-data row except \emph{w/o EL}.}
  \label{tab:ablation}
  \begin{tabular}{lcccccc}
    \toprule
    Training data & TL & PM & SR & OSR & $C$ & $\Delta$SR \\
    \midrule
    Base VLA & -- & -- & 38 & 59 & 0.64 & $-15$ \\
    Geodesic adaptation & -- & -- & 36 & 43 & 0.84 & $-17$ \\
    \midrule
    \oursrow
    \method & \cmark & \cmark & \best{53} & 57 & 0.93 & ref. \\
    \subrow w/o TL & \xmark & \cmark & 47 & 50 & 0.94 & $-6$ \\
    \subrow w/o PM & \cmark & \xmark & 47 & 53 & 0.89 & $-6$ \\
    \subrow w/o EL & \cmark & \cmark & 47 & 53 & 0.89 & $-6$ \\
    \subrow Geodesic actions & \cmark & \cmark & 46 & 57 & 0.81 & $-7$ \\
    \midrule
    Self-imitation & -- & \cmark & 47 & 52 & 0.90 & $-6$ \\
    \bottomrule
  \end{tabular}
\end{table}

Table~\ref{tab:ablation} removes one rule of \method at a time from the model of Tab.~\ref{tab:recipe} and keeps
all other training settings. \emph{Without task-level labels (TL)}, the VLA learns only from steps it
made itself, under the instruction it received at that step. \emph{Without progress masking (PM)}, every
step enters the loss. \emph{Without end labels (EL)}, agent sessions keep an end label only at the pose
where the robot stopped. \emph{Geodesic actions} keeps the states and end labels of the sessions but
replaces every motion target with the geodesic expert's target $y^{\mathrm{geo}}_t$ at $p_t$. These four
variants change Eqs.~\ref{eq:rc} to~\ref{eq:term} to
\begin{equation}
  \begin{aligned}
    &\text{w/o TL} &&\{(o_{\le t},\ell_t,y_t) \mid c_t=\cexec\},\\
    &\text{w/o PM} &&m_t=1 \ \text{for every } t,\\
    &\text{w/o EL} &&m_t=0 \ \text{if } y_t=\mathbf{0} \text{ and } p_t\neq p_T,\\
    &\text{Geodesic actions} &&y_t=y^{\mathrm{geo}}_t \ \text{for every } y_t\neq\mathbf{0}.
  \end{aligned}
  \label{eq:ablations}
\end{equation}
\emph{Self-imitation} trains the VLA on its own successful runs on the same training episodes.

Every rule adds to the success rate. Removing task-level labels, progress masking or the end labels on
agent states lowers SR from 53\% to 47\%, and geodesic action targets lower it to 46\%. Self-imitation
falls 6\% short, so success filtering alone does not explain the gain. The agent's states help too, as
geodesic action targets beat geodesic adaptation by 10\%.

\section{Conclusion}
\label{sec:conclusion}
\reviewlastpage
We showed that a VLA can learn the skills of the LLM agent that calls it and then navigate alone.
Trained with \method, the VLA learns to stop at the goal, reach distant goals and recover after it goes
off track. It sets a new state of the art on \reverie and NavRAG-CE and raises success rate by 15.6\% and
11.1\% on them. It also beats the success rate of its agent on unseen episodes with no LLM tokens. We
study one VLA with one LLM agent; broader instructions and backbones are natural extensions. Running
alone at the step cost of the base VLA, \method makes on-device deployment viable. Code, adapters and agent
sessions will be released upon acceptance.

\FloatBarrier
\clearpage
\section*{LLM Usage Statement}
The agent studied in this paper uses the LLM gpt-6-astra as its planner. For the preparation of the paper
itself, the authors designed every research idea, hypothesis, claim and conclusion, and used an AI tool
only where listed below. The authors checked every result, citation, figure and line of code, and take
full responsibility for all content.

\begin{itemize}
  \item \textbf{Research ideas, hypotheses, claims or conclusions.} None. Every idea, hypothesis, claim
  and conclusion in this paper originates with the authors.
  \item \textbf{Experiments: design, execution, analysis of results.} None. The authors designed the
  protocol, ran every experiment, and read every result.
  \item \textbf{Data: synthetic data, dataset creation, annotation.} None. All agent sessions and
  evaluation runs are collected by the authors' own pipeline from public benchmarks.
  \item \textbf{Code and implementation.} Generate. Claude Code and Codex produced drafts of training
  and evaluation scripts under the authors' prompts; the authors reviewed, edited and tested the
  resulting code before every experiment.
  \item \textbf{Writing and editing text.} Assist. The authors drafted every section; an AI tool was
  used only to refine wording and fix grammar on author-produced prose.
  \item \textbf{Figures, diagrams, visual material.} Assist. The authors chose every figure, provided
  the data and the layout, and used an AI tool to refine style on author-produced figures.
  \item \textbf{Related work and citations.} Assist. The authors selected every reference. An AI tool
  helped align sentences with these references; the authors verified titles, venues and years.
\end{itemize}

{
    \small
    \bibliographystyle{ieeenat_fullname}
    \bibliography{refs}

\begin{thebibliography}{40}
\providecommand{\natexlab}[1]{#1}
\providecommand{\url}[1]{\texttt{#1}}
\expandafter\ifx\csname urlstyle\endcsname\relax
  \providecommand{\doi}[1]{doi: #1}\else
  \providecommand{\doi}{doi: \begingroup \urlstyle{rm}\Url}\fi

\bibitem[Anderson et~al.(2018{\natexlab{a}})Anderson, Chang, Chaplot,
  et~al.]{anderson2018eval}
Peter Anderson, Angel Chang, Devendra~Singh Chaplot, et~al.
\newblock On evaluation of embodied navigation agents.
\newblock \emph{arXiv preprint arXiv:1807.06757}, 2018{\natexlab{a}}.

\bibitem[Anderson et~al.(2018{\natexlab{b}})Anderson, Wu, Teney,
  et~al.]{anderson2018r2r}
Peter Anderson, Qi Wu, Damien Teney, et~al.
\newblock Vision-and-language navigation: Interpreting visually-grounded
  navigation instructions in real environments.
\newblock In \emph{CVPR}, 2018{\natexlab{b}}.

\bibitem[Bonetta et~al.(2026)Bonetta, Merler, Zago, Cancelliere, and
  Magnini]{sage}
Giovanni Bonetta, Matteo Merler, Davide Zago, Rossella Cancelliere, and
  Bernardo Magnini.
\newblock Selective agent guidance via entropy: Learning autonomous policies
  from imperfect {VLM} teachers.
\newblock In \emph{Findings of EMNLP}, 2026.

\bibitem[Chang et~al.(2017)Chang, Dai, Funkhouser, et~al.]{chang2017mp3d}
Angel Chang, Angela Dai, Thomas Funkhouser, et~al.
\newblock {Matterport3D}: Learning from {RGB-D} data in indoor environments.
\newblock In \emph{3DV}, 2017.

\bibitem[Cheng et~al.(2025)Cheng, Ji, Yang, et~al.]{navila}
An-Chieh Cheng, Yandong Ji, Zhaojing Yang, et~al.
\newblock {NaVILA}: Legged robot vision-language-action model for navigation.
\newblock In \emph{RSS}, 2025.

\bibitem[Choudhury and Sodhi(2025)]{leap}
Sanjiban Choudhury and Paloma Sodhi.
\newblock Better than your teacher: {LLM} agents that learn from privileged
  {AI} feedback.
\newblock In \emph{ICLR}, 2025.

\bibitem[Gong et~al.(2026)Gong, Zheng, Hou, et~al.]{tinyastral}
Kehong Gong, Xiaoxu Zheng, Guanli Hou, et~al.
\newblock {TinyAstral}: Transferring self-evolving harnesses to tiny generalist
  manipulation agents.
\newblock \emph{arXiv preprint arXiv:2610.41216}, 2026.

\bibitem[He et~al.(2026)He, Liu, Xu, et~al.]{budvln}
Gang He, Zhenyang Liu, Kepeng Xu, et~al.
\newblock Nipping the drift in the bud: Retrospective rectification for robust
  vision-language navigation.
\newblock \emph{arXiv preprint arXiv:2602.06356}, 2026.

\bibitem[Hu et~al.(2022)Hu, Shen, Wallis, Allen-Zhu, Li, Wang, Wang, and
  Chen]{hu2022lora}
Edward~J. Hu, Yelong Shen, Phillip Wallis, Zeyuan Allen-Zhu, Yuanzhi Li, Shean
  Wang, Lu Wang, and Weizhu Chen.
\newblock {LoRA}: Low-rank adaptation of large language models.
\newblock In \emph{ICLR}, 2022.

\bibitem[Ju et~al.(2026)Ju, Cai, Xu, et~al.]{xfly}
Ruibin Ju, Ruifeng Cai, Min Xu, et~al.
\newblock {X-Fly}: Learning to reach and stop in full-chunk aerial
  vision-language navigation.
\newblock \emph{arXiv preprint arXiv:2610.21450}, 2026.

\bibitem[Kim et~al.(2026)Kim, Kim, and Lee]{phinav}
Sung~June Kim, Sangpil Kim, and Honglak Lee.
\newblock Path-level hindsight instructions for semantic exploration in
  vision-language navigation.
\newblock In \emph{ECCV}, 2026.

\bibitem[Krantz et~al.(2020)Krantz, Wijmans, Majumdar, Batra, and
  Lee]{krantz2020vlnce}
Jacob Krantz, Erik Wijmans, Arjun Majumdar, Dhruv Batra, and Stefan Lee.
\newblock Beyond the nav-graph: Vision-and-language navigation in continuous
  environments.
\newblock In \emph{ECCV}, 2020.

\bibitem[Laskey et~al.(2017)Laskey, Lee, Fox, Dragan, and
  Goldberg]{laskey2017dart}
Michael Laskey, Jonathan Lee, Roy Fox, Anca Dragan, and Ken Goldberg.
\newblock {DART}: Noise injection for robust imitation learning.
\newblock In \emph{CoRL}, 2017.

\bibitem[Lauffer et~al.(2026)Lauffer, Deng, Kundurthy, Kenstler, and Da]{oec}
Niklas Lauffer, Xiang Deng, Srivatsa Kundurthy, Brad Kenstler, and Jeff Da.
\newblock Imitation learning for multi-turn {LM} agents via on-policy expert
  corrections.
\newblock In \emph{ICML}, 2026.

\bibitem[Lei et~al.(2026)Lei, Zhou, Chen, et~al.]{navmcp}
Zixing Lei, Gengze Zhou, Xiong-Hui Chen, et~al.
\newblock Scaffolding foundation models into physical-world agents pushes the
  frontier of long-horizon navigation.
\newblock \emph{arXiv preprint arXiv:2608.30396}, 2026.

\bibitem[Li et~al.(2025)Li, Wang, Zhou, et~al.]{sid}
Songze Li, Zun Wang, Gengze Zhou, et~al.
\newblock Learning goal-oriented vision-and-language navigation with
  self-improving demonstrations at scale.
\newblock \emph{arXiv preprint arXiv:2509.24910}, 2025.

\bibitem[Li et~al.(2026)Li, Wu, Wang, et~al.]{hsl}
Zichao Li, Gang Wu, Zichao Wang, et~al.
\newblock Spinning straw into gold: Relabeling {LLM} agent trajectories in
  hindsight for successful demonstrations.
\newblock In \emph{ICLR}, 2026.

\bibitem[Liu et~al.(2026{\natexlab{a}})Liu, Li, Yao, et~al.]{guava}
Haowen Liu, Xirui Li, Shaoxiong Yao, et~al.
\newblock Guava: Distilling frontier {VLMs} into a compact agent through a
  robotic manipulation harness.
\newblock \emph{arXiv preprint arXiv:2606.18363}, 2026{\natexlab{a}}.

\bibitem[Liu et~al.(2026{\natexlab{b}})Liu, Tian, Qiu, et~al.]{stopwatch}
Xinhao Liu, Yunsheng Tian, Yuheng Qiu, et~al.
\newblock {StopWatch}: Goal grounding for object-goal navigation.
\newblock \emph{arXiv preprint arXiv:2610.23841}, 2026{\natexlab{b}}.

\bibitem[Long et~al.(2024)Long, Cai, Wang, Zhan, and Dong]{instructnav}
Yuxing Long, Wenzhe Cai, Hongcheng Wang, Guanqi Zhan, and Hao Dong.
\newblock {InstructNav}: Zero-shot system for generic instruction navigation in
  unexplored environment.
\newblock In \emph{CoRL}, 2024.

\bibitem[Qi et~al.(2020)Qi, Wu, Anderson, Wang, Wang, Shen, and van~den
  Hengel]{qi2020reverie}
Yuankai Qi, Qi Wu, Peter Anderson, Xin Wang, William~Yang Wang, Chunhua Shen,
  and Anton van~den Hengel.
\newblock {REVERIE}: Remote embodied visual referring expression in real indoor
  environments.
\newblock In \emph{CVPR}, 2020.

\bibitem[Ramrakhya et~al.(2022)Ramrakhya, Undersander, Batra, and
  Das]{habitatweb}
Ram Ramrakhya, Eric Undersander, Dhruv Batra, and Abhishek Das.
\newblock {Habitat-Web}: Learning embodied object-search strategies from human
  demonstrations at scale.
\newblock In \emph{CVPR}, 2022.

\bibitem[Ramrakhya et~al.(2023)Ramrakhya, Batra, Wijmans, and Das]{pirlnav}
Ram Ramrakhya, Dhruv Batra, Erik Wijmans, and Abhishek Das.
\newblock {PIRLNav}: Pretraining with imitation and {RL} finetuning for
  {ObjectNav}.
\newblock In \emph{CVPR}, 2023.

\bibitem[Savva et~al.(2019)Savva, Kadian, Maksymets, et~al.]{savva2019habitat}
Manolis Savva, Abhishek Kadian, Oleksandr Maksymets, et~al.
\newblock Habitat: A platform for embodied {AI} research.
\newblock In \emph{ICCV}, 2019.

\bibitem[Shi et~al.(2026)Shi, Yang, Tao, et~al.]{realm}
Xiangyu Shi, Ruoxi Yang, Wei Tao, et~al.
\newblock From region arrival to instance-level grounding in
  vision-and-language navigation.
\newblock \emph{arXiv preprint arXiv:2607.03792}, 2026.

\bibitem[Tang et~al.(2026)Tang, Allen, van Steenkiste, et~al.]{decoupleplan}
Zineng Tang, Kelsey~R. Allen, Sjoerd van Steenkiste, et~al.
\newblock Decoupling planning and control for instructable agents.
\newblock In \emph{COLM}, 2026.

\bibitem[Wang and Lee(2025)]{g3dlf}
Zihan Wang and Gim~Hee Lee.
\newblock {g3D-LF}: Generalizable {3D}-language feature fields for embodied
  tasks.
\newblock In \emph{CVPR}, 2025.

\bibitem[Wang et~al.(2025{\natexlab{a}})Wang, Lee, and Lee]{wang2025dynam3d}
Zihan Wang, Seungjun Lee, and Gim~Hee Lee.
\newblock {Dynam3D}: Dynamic layered {3D} tokens empower {VLM} for
  vision-and-language navigation.
\newblock In \emph{NeurIPS}, 2025{\natexlab{a}}.

\bibitem[Wang et~al.(2025{\natexlab{b}})Wang, Li, Hong, et~al.]{srdf}
Zun Wang, Jialu Li, Yicong Hong, et~al.
\newblock Bootstrapping language-guided navigation learning with self-refining
  data flywheel.
\newblock In \emph{ICLR}, 2025{\natexlab{b}}.

\bibitem[Wang et~al.(2025{\natexlab{c}})Wang, Zhu, Lee, and
  Fan]{yang2024navrag}
Zihan Wang, Yaohui Zhu, Gim~Hee Lee, and Yachun Fan.
\newblock {NavRAG}: Generating user demand instructions for embodied navigation
  through retrieval-augmented {LLM}.
\newblock In \emph{Findings of ACL}, 2025{\natexlab{c}}.

\bibitem[Wang et~al.(2026)Wang, Lee, Dai, and Lee]{d3dvlp}
Zihan Wang, Seungjun Lee, Guangzhao Dai, and Gim~Hee Lee.
\newblock {D3D-VLP}: Dynamic {3D} vision-language-planning model for embodied
  grounding and navigation.
\newblock In \emph{CVPR}, 2026.

\bibitem[Wei et~al.(2026{\natexlab{a}})Wei, Wan, Peng, et~al.]{dualvln}
Meng Wei, Chenyang Wan, Jiaqi Peng, et~al.
\newblock Ground slow, move fast: A dual-system foundation model for
  generalizable vision-language navigation.
\newblock In \emph{ICLR}, 2026{\natexlab{a}}.

\bibitem[Wei et~al.(2026{\natexlab{b}})Wei, Wan, Yu, et~al.]{wei2025streamvln}
Meng Wei, Chenyang Wan, Xiqian Yu, et~al.
\newblock {StreamVLN}: Streaming vision-and-language navigation via {SlowFast}
  context modeling.
\newblock In \emph{ICRA}, 2026{\natexlab{b}}.

\bibitem[{Xiaomi Embodied Intelligence Team}(2026)]{omnibrain0}
{Xiaomi Embodied Intelligence Team}.
\newblock {OmniBrain-0}: A unified agentic model for embodied intelligence.
\newblock \emph{arXiv preprint arXiv:2610.36013}, 2026.

\bibitem[Ye et~al.(2026{\natexlab{a}})Ye, Lu, Dong, et~al.]{harnesszero}
Haoran Ye, Yuxing Lu, Haonan Dong, et~al.
\newblock {Harness-Zero}: Harness distillation via agent-as-harness.
\newblock \emph{arXiv preprint arXiv:2609.24974}, 2026{\natexlab{a}}.

\bibitem[Ye et~al.(2026{\natexlab{b}})Ye, Wei, Wang, et~al.]{match}
Shuhao Ye, Yufei Wei, Demao Wang, et~al.
\newblock {MATCH}: Execution-aware hierarchical reinforcement learning for
  vision-and-language navigation in continuous environment.
\newblock \emph{arXiv preprint arXiv:2610.38248}, 2026{\natexlab{b}}.

\bibitem[Zhang et~al.(2024)Zhang, Wang, Xu, et~al.]{navid}
Jiazhao Zhang, Kunyu Wang, Rongtao Xu, et~al.
\newblock {NaVid}: Video-based {VLM} plans the next step for
  vision-and-language navigation.
\newblock In \emph{RSS}, 2024.

\bibitem[Zhang et~al.(2025)Zhang, Wang, Wang, et~al.]{uninavid}
Jiazhao Zhang, Kunyu Wang, Shaoan Wang, et~al.
\newblock {Uni-NaVid}: A video-based vision-language-action model for unifying
  embodied navigation tasks.
\newblock In \emph{RSS}, 2025.

\bibitem[Zhang et~al.(2026)Zhang, Zhou, Yin, et~al.]{qwenrobotnav}
Jiazhao Zhang, Gengze Zhou, Hale Yin, et~al.
\newblock {Qwen-RobotNav} technical report: A scalable navigation model
  designed for an agentic navigation system.
\newblock \emph{arXiv preprint arXiv:2606.18112}, 2026.

\bibitem[Zhou et~al.(2026)Zhou, Hong, Zhang, et~al.]{navgpt3}
Gengze Zhou, Yicong Hong, Jiazhao Zhang, et~al.
\newblock {NavGPT-3}: Harnessing context in a hierarchical navigation runtime.
\newblock \emph{arXiv preprint arXiv:2610.10787}, 2026.

\end{thebibliography}
}

\ifdefined\cvprsubmit\else

\fi
\fi

\end{document}